%% file: neurips_data_2023.tex
\documentclass{article}

\usepackage[preprint]{neurips_data_2023}

\input{packages}
\input{macros}

\title{
\includegraphics[height=.4cm]{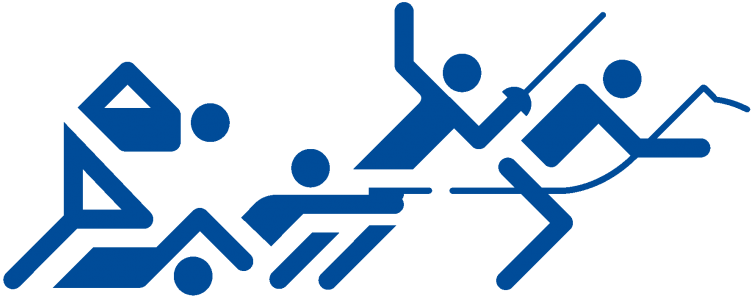}
Efficiency Pentathlon: \\
A Standardized Arena for Efficiency Evaluation
}
\author{%
\textbf{Hao Peng}$^\spadesuit$ \quad
    \textbf{Qingqing Cao}$^\diamondsuit$ \quad
    \textbf{Jesse Dodge}$^\spadesuit$ \quad
    \textbf{Matthew E. Peters}$^\spadesuit$ \quad
    \textbf{Jared Fernandez}$^\heartsuit$ \quad \\
    \textbf{Tom Sherborne}$^{\clubsuit}$\thanks{\hspace{1mm}This work was done during Tom Sherborne's internship  at AI2. $\dagger$ denotes equal contribution.}\quad
    \textbf{Kyle Lo}$^{\spadesuit\dagger}$ \quad
    \textbf{Sam Skjonsberg}$^{\spadesuit\dagger}$ \quad
    \textbf{Emma Strubell}$^{\heartsuit\spadesuit\dagger}$ \quad \\
    \textbf{Darrell Plessas}$^{\spadesuit\dagger}$ \quad
    \textbf{Iz Beltagy}$^{\spadesuit\dagger}$ \quad
    \textbf{Evan Pete Walsh}$^{\spadesuit\dagger}$ \quad \\
    \textbf{Noah A. Smith}$^{\diamondsuit\spadesuit}$\quad
    \textbf{Hannaneh Hajishirzi}$^{\diamondsuit\spadesuit}$ \quad\\
     $^\spadesuit$Allen Institute for Artificial Intelligence\\ 
  $^\diamondsuit$Paul G. Allen School of Computer Science \& Engineering,
  University of Washington\\
  $^\heartsuit$Language Technologies Institute, Carnegie Mellon University\\
  $^\clubsuit$Institute for Language, Cognition and Computation, University of Edinburgh\\
  {\tt \{haop,jessed,matthewp,kylel,sams,darrellp,beltagy,petew\}@allenai.org} \\
  {\tt \{qicao,nasmith,hannaneh\}@cs.washington.edu}\\
  {\tt \{jaredfern,strubell\}@cmu.edu},
  {\tt tom.sherborne@ed.ac.uk}
}
\begin{document}

\maketitle
\input{text/abstract}

\input{text/intro}
\input{text/platform}
\input{text/experiments}
\input{text/related}
\input{text/conclusion}

\bibliography{neurips_data_2023}
\bibliographystyle{neurips_data_2023}

\clearpage
\input{text/checklist}

\clearpage
\appendix
\input{text/appendix}

\end{document}

%% file: packages.tex
\usepackage{neurips_data_2023}
\usepackage[utf8]{inputenc} 
\usepackage[T1]{fontenc}    
\usepackage{hyperref}       
\usepackage{url}            
\usepackage{booktabs}       
\usepackage{amsfonts}       
\usepackage{nicefrac}       
\usepackage{microtype}      
\usepackage{xcolor}         
\usepackage{standalone}
\usepackage{latexsym}
\usepackage{amsmath}
\usepackage{amssymb}
\usepackage{amsthm}
\usepackage{graphicx}
\usepackage{subcaption}
\usepackage{array}
\usepackage{tabu}
\usepackage{makecell}
\usepackage{paralist}
\usepackage{cases}
\usepackage{diagbox}
\usepackage{enumitem}
\usepackage{soul}
\usepackage{multirow}
\usepackage{verbatim}
\usepackage{tabulary}
\usepackage{booktabs}
\usepackage[mathscr]{euscript}
\usepackage{mathtools}
\usepackage{algorithm}
\usepackage{algpseudocode}
\usepackage{stmaryrd}
\usepackage{tikz-dependency}
\usetikzlibrary{automata,decorations.markings,arrows,positioning,matrix,calc,patterns,angles,quotes,calc}
\usepackage{adjustbox}
\usepackage{tabularx}
\usepackage{xspace}
\usepackage{tabulary}
\usepackage{afterpage}
\usepackage{bm}
\usepackage{color}
\usepackage{graphicx}
\usepackage{slashbox}
\usepackage[toc,page]{appendix}
\usepackage{makecell}
\usepackage{boldline}
\usepackage[shortcuts]{extdash}  

\usepackage{blindtext}
\usepackage{graphicx}
\usepackage{capt-of}
\usepackage{booktabs}
\usepackage{varwidth}
\usepackage{pifont}
\usepackage{wrapfig}

\usepackage{listings}

\usepackage{pythonhighlight}

%% file: macros.tex
\definecolor{orange}{rgb}{1,0.5,0}
\definecolor{mdgreen}{rgb}{0.05,0.6,0.05}
\definecolor{mdblue}{rgb}{0,0,0.7}
\definecolor{dkblue}{rgb}{0,0,0.5}
\definecolor{dkgray}{rgb}{0.3,0.3,0.3}
\definecolor{slate}{rgb}{0.25,0.25,0.4}
\definecolor{gray}{rgb}{0.5,0.5,0.5}
\definecolor{ltgray}{rgb}{0.7,0.7,0.7}
\definecolor{purple}{rgb}{0.7,0,1.0}
\definecolor{lavender}{rgb}{0.65,0.55,1.0}

\definecolor{mypurple}{RGB}{111,61,121}
\definecolor{myblue}{RGB}{46,88,180}
\definecolor{myred}{RGB}{181,68,106}
\definecolor{myyellow}{RGB}{204,143,55}

\newcommand{\ensuretext}[1]{#1}
\newcommand{\marker}[2]{\ensuremath{^{\textsc{#1}}_{\textsc{#2}}}}
\newcommand{\draftcomment}[3]{\ensuretext{\textcolor{#3}{[#1 #2]}}}
\renewcommand{\draftcomment}[3]{}  

\newcommand{\matt}[1]{\draftcomment{\marker{M}{P}}{#1}{olive}}

\newcommand{\kyle}[1]{\draftcomment{\marker{K}{L}}{#1}{lavender}}

\newcommand{\cmark}{\ding{51}}%
\newcommand{\xmark}{\ding{55}}%

\newcommand{\interalia}[1]{\citep[\emph{inter alia}]{#1}}

\DeclareSymbolFont{extraup}{U}{zavm}{m}{n}
\DeclareMathSymbol{\vardiamond}{\mathalpha}{extraup}{87}

\newcolumntype{L}[1]{>{\raggedright\let\newline\\\arraybackslash\hspace{0pt}}m{#1}}
\newcolumntype{C}[1]{>{\centering\let\newline\\\arraybackslash\hspace{0pt}}m{#1}}
\newcolumntype{R}[1]{>{\raggedleft\let\newline\\\arraybackslash\hspace{0pt}}m{#1}}

\theoremstyle{definition}

\theoremstyle{remark}

\algrenewcommand{\algorithmiccomment}[1]{\leavevmode$\triangleright$ #1}

\setul{1pt}{.4pt}

\newcommand{\name}{Pentathlon\xspace}

\newcommand{\stdin}{\texttt{stdin}\xspace}
\newcommand{\stdout}{\texttt{stdout}\xspace}
\newcommand{\stdio}{\texttt{stdio}\xspace}
\newcommand{\wmt}{WMT14 DE-EN\xspace}

\DeclareFixedFont{\ttb}{T1}{txtt}{bx}{n}{12} 
\DeclareFixedFont{\ttm}{T1}{txtt}{m}{n}{12}  

%% file: text/abstract.tex
\begin{abstract}
Rising computational demands of modern natural language processing (NLP) systems have increased the barrier to entry for cutting-edge research while posing serious environmental concerns. Yet, progress on model efficiency has been impeded by practical challenges in model evaluation and comparison. For example, hardware is challenging to control due to disparate levels of accessibility across different institutions. Moreover, improvements in metrics such as FLOPs often fail to translate to progress in real-world applications.
In response, we introduce efficiency \name, a benchmark for holistic and realistic evaluation of model efficiency. \name focuses on inference, which accounts for a majority of the compute in a model’s lifecycle. It offers a strictly-controlled hardware platform, and is designed to mirror real-world applications scenarios. It incorporates a suite of metrics that target different aspects of efficiency, including latency, throughput, memory overhead, number of parameters, and energy consumption, hence the name {\bf Penta}thlon. 
It also comes with a software library that can be seamlessly integrated into any codebase and enable evaluation. 
As a standardized and centralized evaluation platform, \name can drastically reduce the workload to make fair and reproducible efficiency comparisons. 
While initially focused on natural language processing (NLP) models, \name is designed to allow flexible extension to other fields.
We envision \name will stimulate algorithmic innovations in building efficient models, and foster an increased awareness of the social and environmental implications in the development of future-generation NLP models.

\end{abstract}

%% file: text/intro.tex

\section{Introduction \label{sec:intro}}
The remarkable recent progress in artificial intelligence owes much to advances in large-scale deep learning models~\interalia{brown2020gpt,chowdhery2022palm,llama}.
However, their rapidly-increasing computational demands have introduced substantial challenges. 
The barrier to entry to cutting-edge research is raised, particularly impacting researchers and practitioners with fewer resources and exacerbating disparities in the AI research landscape.
Moreover, the escalating energy consumption associated with these computation-intensive models leads to serious environmental concerns~\interalia{Lacoste2019QuantifyingTC,schwartz2020green,henderson2020towards,Strubell2020EnergyAP}.

Therefore, building more efficient models for AI systems has become a pressing challenge, drawing widespread attention from the community~\interalia{2020-mlperf-reddi,tay2020efficient,treviso2022efficient,liu-etal-2022-towards-efficient,Yao2022ZeroQuantEA,fu2023specializing}. However, a lack of standardized evaluation protocols makes it challenging to measure the progress in efficiency improvements and obstructs the efforts in developing more efficient models. In many cases, models are evaluated in scenarios that hardly reflect the deployment of machine learning models in practice \citep{henderson2020towards}. Moreover, some widely-adopted efficiency metrics such as FLOPs often poorly correlate with models' real-world efficiency performance \citep{dehghani2022the,fernandez2023framework}.
\begin{wrapfigure}[20]{r}{0.5\textwidth}
\includegraphics[width=0.5\textwidth, trim={15cm 10cm 12.5cm 10.5cm},clip]{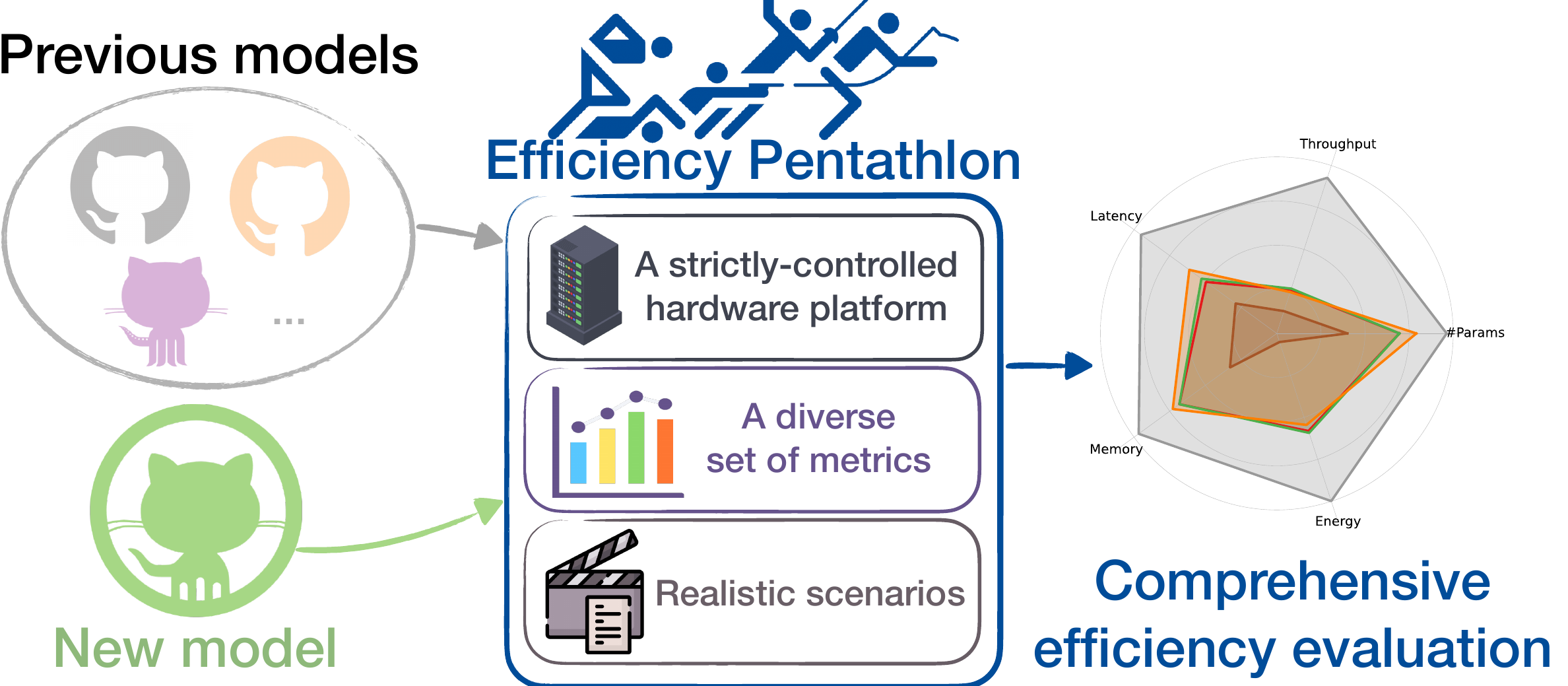}
\caption{By submitting to \name, practitioners can compare their models against all previous submissions on identical hardware, eliminating the need to re-implement previous works and substantially reducing the workloads for fair efficiency comparisons.
Models are evaluated in four realistic scenarios designed to mirror real-world applications.
Our platform evaluates the submission across five crucial efficiency metrics, including throughput, latency, memory overhead, the number of parameters, and energy consumption, hence the name {\bf Penta}thlon.
}\label{fig:diagram}
\end{wrapfigure}
The issue is exacerbated by several practical challenges.  For instance, hardware is a critical confounding factor in efficiency comparisons, but is very challenging to control in practice, due to disparate levels of hardware accessibility across institutions. Consequently, this leads to disconnections between efficiency improvements in research and tangible progress in practice. There is a pressing need for a standardized efficiency evaluation framework. 


To address these challenges, we present \name. 
It is designed to establish a standardized platform for evaluating the \emph{inference} efficiency of AI models. As shown by \citet{patterson22} and \citet{wu2022sustainable}, inference accounts for over 60\% of energy consumption in real-world machine learning workloads. 
\name aims to provide comprehensive and realistic evaluation of efficiency, and offer the community a platform to make fair comparisons in a strictly controlled environment. 
To achieve this, we make several key design choices:
\begin{itemize}[leftmargin=*,noitemsep,topsep=0pt,parsep=0pt,partopsep=0pt]
\item \textbf{Controlled hardware environment} (\S\ref{sec:controlling}): hosted by a dedicated server, \name provides a centralized platform with a strictly controlled hardware environment. This removes the necessity for practitioners to reproduce previous works on their own hardware for fair comparisons and allows easy comparisons with models previously evaluated on \name using identical hardware.
Moreover, it allows us to use power monitoring devices to accurately measure the energy consumption during models' inference, which was previously impossible.
\item \textbf{Realistic scenarios} (\S\ref{sec:realistic}):  It evaluates models under various scenarios specifically designed to mirror real-world deployment contexts, allowing different approaches to batching input instances, aiming to bridge the gap between research context and practical applications.
\item \textbf{Comprehensive metrics} (\S\ref{sec:diverse}): \name evaluates models with five crucial metrics, including throughput, latency, memory overhead, the number of parameters, and energy consumption, hence the name {\bf Penta}thlon.  This provides a more holistic understanding of a model's efficiency.
\item \textbf{Flexibility} (\S\ref{sec:flexibility}) \name is flexible by design and can be seamlessly integrated into any codebase. Although we focus on natural language processing (NLP) models in this paper, \name can be easily extended to other fields.
\end{itemize}

\name is ready to accept submissions, helping to reduce the workload of conducting fair efficiency comparisons: \url{https://github.com/allenai/efficiency-pentathlon}.
As we demonstrate in the experiments~(\S\ref{sec:experiments}), \name can provide fresh insights into existing models.
Through our comparisons of several established machine translation models,
the comprehensive evaluation offered by \name highlights the particular effectiveness of quantization in large models.
Furthermore, \name's energy evaluation component reveals new perspectives on the models' energy consumption during inference.

We envision that by offering standardized efficiency evaluation, \name will stimulate the development of more efficient models and foster a deeper awareness of the computational costs of AI research, and accelerate progress on reducing them. 

%% file: text/platform.tex
\section{\includegraphics[height=.3cm]{fig/logo} Efficiency \name}\label{sec:efficiency_benchmark}


This section discusses the current challenges in efficiency evaluation and outlines the design choices we adopted in \name to effectively address them.

\subsection{Controlling the Hardware for Fair Efficiency Comparisons} \label{sec:controlling}

The hardware stands as a critical confounding factor when comparing efficiency,
and can significantly influence the conclusions of such comparisons. 
As demonstrated by several recent studies, the trends in efficiency comparisons can vary substantially when different accelerators are used \interalia{peng2021rfa,kasai2021finetuning,wu2022modeling, wang2020systematic}.
Compounding this issue is the practical difficulty in controlling for hardware, primarily because access to hardware platforms often varies among institutions.
This is a major obstacle for fair efficiency comparisons.
Even with publicly available implementations, practitioners often need to adapt these to their own hardware environments to ensure fair comparisons.

\noindent \textbf{Our approach.} 
\name aims to stimulate algorithmic innovations that can generalize across different hardware.
Therefore we control for hardware while conducting efficiency comparisons 
and offer a varied selection of hardware to simulate different use cases.
\name is hosted with a dedicated in-house server.
Participants can submit their models' code and checkpoints to our server through an easy-to-use tool that we provide (\S\ref{sec:flexibility}).
This ensures that all models evaluated using \name use an identical hardware environment, guaranteeing  fair comparisons.
By requiring code submission \name helps improve transparency.
The specific implementation choices for each submission, such as data IO and padding, will be thoroughly documented.
This is appealing because it  helps disentangle the efficiency gains due to \emph{algorithmic innovations} from those achieved by better implementations that can equally benefit all models.
Further, a dedicated in-house server allows us to measure energy consumption, which would otherwise be very challenging to incorporate (\S\ref{sec:diverse}). 

The hosting machine of \name has two NVIDIA RTX 8000 GPUs,
two Intel Xeon Ice Lake Gold 6348 28-Core CPUs, and 1TB DDR4 memory.
It supports evaluation using GPUs and CPUs, and CPUs only.
We plan to extend \name to offer a broader selection of hardware in the near future.\footnote{
    We plan to use the NVIDIA Jetson TX2 Module (\url{https://developer.nvidia.com/embedded/jetson-tx2}) to simulate limited-resource settings such as on an automobile,
    and extend \name to a smartphone to evaluate machine learning models designed to run on mobile devices.
}

To accurately measure each submission's efficiency without interference, we have implemented a scheduler on the server. This ensures that only one inference workload is running at any given time. 
In \name, the efficiency measurement begins when the model has been loaded and is ready for predictions, excluding the overhead associated with both model and data loading. 

\subsection{Realistic Evaluation Scenarios Designed to Emulate Real-world Applications} \label{sec:realistic}
\input{tables/scenarios}
NLP systems are deployed across a broad range of practical applications, each with its unique requirements for efficiency.
Consider, for instance, an online search engine. 
The arrivals of users' queries are unpredictable, and so is the model's inference batch size.
An AI assistant operating on a smartphone typically processes one request at a time,
while an offline translation system translating an entire book must use large batch sizes to prioritize maximizing throughput.
These practical scenarios are rarely reflected by conventional efficiency evaluations in the research context, where models are typically assessed with a fixed batch size. Such disparity underscores the pressing need for evaluation protocols that better reflect real-world deployments.

\noindent \textbf{Our approach.}
Inspired by \citet{2020-mlperf-reddi}, we include four distinct evaluation scenarios to provide a comprehensive evaluation of NLP models in a variety of realistic settings:
\begin{itemize}[leftmargin=*,noitemsep,topsep=0pt,parsep=0pt,partopsep=0pt]
\item {\bf Fixed batching.} The evaluation data is first randomly shuffled before being grouped into batches of a user-specified \texttt{batch-size}. This setting is intended to mimic typical research experimental settings. We  defer to the users choosing optimal batch sizes for their models.
\item {\bf Poisson batching} is similar to the fixed batching scenario, but the size of each batch is randomly drawn from a Poisson distribution with a mean of \texttt{batch-size}: $\texttt{batch-size}_{\text{Pois}} \sim \text{Pois}(\texttt{batch-size})$. This setup aims to simulate an online service where the volume of requests is unpredictable but the average can be estimated.
\item {\bf Single stream} randomly shuffles the evaluation instances and uses a batch size of one, reflecting the applications processing one request at a time.
\item {\bf Offline:} In this scenario, the model has immediate access to the entire evaluation dataset, enabling techniques such as sorting the inputs by length or adaptive batching to enhance throughput and memory efficiency. This scenario reflects large-scale, offline tasks.
\end{itemize}
These varied evaluation scenarios are designed to highlight the strengths and weaknesses of different models in diverse deployment contexts.
\subsection{A Diverse Set of Metrics for Comprehensive Efficiency Evaluation} \label{sec:diverse}
AI systems' efficiency in practical contexts is multifaceted and can hardly be adequately represented by any single metric.
Different use cases prioritize different efficiency aspects.
For example, a model deployed on mobile devices prioritizes energy efficiency, an offline model requires optimal throughput, while an online service model demands low latency.
However, the widely-used metrics often fail to show strong correlations with these diverse practical aspects of efficiency.
Take, for instance, the number of floating point number operations (FLOPs) a model takes for performing a workload.
It has become a standard efficiency metric partly due to its hardware and implementation-agnostic nature, highlighting the algorithmic advancements in model efficiency \citep{schwartz2020green}. 
Yet recent research has cast doubt on its relevance, showing that it is a poor indicator of many practical metrics including throughput, latency, and energy consumption~\citep{henderson2020towards}.
Even for models sharing similar architectures and numbers of parameters, 
their energy efficiency can diverge significantly under identical workloads, partly due to specific deep learning operations they are implemented with \citep{cao2021irene}.

This highlights the limitations of conventional evaluation protocols, which risk oversimplifying efficiency comparisons by attempting to encapsulate performance in a single measure. 
Instead, we propose a more comprehensive approach that considers a diverse suite of metrics. 
It more accurately reflects the multifaceted nature of efficiency in AI models.

\noindent \textbf{Our approach.}
Our benchmark's  suite of evaluation metrics includes the following:
\begin{itemize}[leftmargin=*,noitemsep,topsep=0pt,parsep=0pt,partopsep=0pt]
    \item {\bf Throughput} measures the volume of data a system can process in a unit of time. 
    We measure throughput with instances/s; for tasks that require generating text, we also consider words/s.
    \item {\bf Latency}, in milliseconds. It quantifies the delay between the system receiving a user request and providing a response.
    Complementing throughput, it's especially critical in real-time applications, such as smartphone-based AI assistants.
    \item {\bf Memory overhead}, in GiB,
    provides insight into a system's applicability in low-resource settings, where available memory can be a bottleneck. 
    In resource-abundant settings, lower memory overhead allows larger batch sizes during inference, improving metrics such as throughput. 
    Our benchmark measures maximum CPU and GPU (if applicable) memory consumption.
    \item {\bf Energy consumption and carbon footprint.} 
    The energy overhead of a system, measured in W$\cdot$h, indicates its suitability for battery-powered devices. 
    Combined with carbon intensity data, it can also assess a model's carbon footprint in terms of the amount of CO$_2$ emissions, providing an environmental impact comparison for models deployed in practice.  We provide more details about measuring energy consumption in \S\ref{sec:energy}.
    \item {\bf Model size}, measured in the number of parameters, serves as an indicator of models' storage overhead, and often correlates with its memory overhead. 
\end{itemize}
Our approach provides a holistic view of model efficiency, with each focusing on specific application contexts,
allowing practitioners to select efficient methods catered to their applications.  



\subsubsection{Challenges in Measuring Energy and our Solution} \label{sec:energy}
While most of the metrics above can be measured with existing tools, accurately measuring energy presents unique challenges, primarily due to the lack of established software for this purpose. 
Although CUDA offers toolkits to measure GPU power, the power usage of CPUs, DRAM, and disks is only accessible on specific types hardware and requires root access \citep{rapl}.

Many existing methods estimate energy consumption for \emph{training} using GPU energy alone \citep{luccioni2022estimating,helm}. 
However, as we will demonstrate in the experiments, this approach is not suitable for our purposes for two primary reasons. 
First, it excludes energy comparisons of models running on CPUs, which our study aims to explore. 
Second, inference tasks by nature entail more frequent data IO interactions, imposing more significant workloads on CPUs, DRAM, disks, etc., compared to training. 
In our experiments, they account for more than 60\% of energy consumption—-a significant increase compared to previous estimates for training \citep{dodge2022measuring}.
Therefore, it is essential to measure not only GPU energy but the total energy consumed by the entire machine accurately.

To this end, we use an energy-monitoring device to measure the power
consumption.\footnote{We use an emonTx V4 for power consumption measurement: \url{https://shop.openenergymonitor.com/single-phase-6-channel-energy-monitoring-emontx-v4/}.}
This data, in conjunction with the model's run time, can be used to calculate the model's energy consumption.
Physically connected to the host machine's power cables, this device's sensors provide accurate real-time power usage data. 
According to the manufacturer, the error rate is $\pm 1.2\%$.

The power consumption is calculated by subtracting the host machine's idling power from the meter reading during an inference run. 
To calculate the carbon emissions, we use the carbon intensity data provided by \citet{codecarbon} based on the geographical location and time of the day. 


\subsection{Ensuring Flexibility in \name} \label{sec:flexibility}
Requiring code and checkpoint submission imposes additional implementation effort from participants, a tradeoff we believe is worthwhile for achieving fair comparisons on a strictly-controlled hardware platform. 
Recognizing from past benchmark efforts that this might discourage practitioners from participating,
we have made a concerted effort to ensure that \name can be easily integrated into existing code bases and to streamline the submission process.\footnote{This is a lesson that some of the authors learned from the NAACL2022 reproducibility track: \url{https://2022.naacl.org/blog/reproducibility-track/}}

\noindent \textbf{Accommodating diverse software frameworks.}
We aim to encourage wide participation and ensure our platform is accessible to practitioners accustomed to various software infrastructures. 
Therefore, \name makes no assumption about the submission's deep learning framework (if a deep learning model is used at all) or the programming language it's implemented in. 
We require that every submission:
(1) Include a GitHub repository containing the code and listing dependencies (this repository does not need to be public);
(2) Interface the model to read inputs from \stdin and write outputs to \stdout;\footnote{We provide a Python tool for this \stdio interaction. Users can implement their own interfaces if they decide to use other programming languages.}
(3) Implement the necessary tools to download the model checkpoint for evaluation.
We provide detailed instructions and examples to guide practitioners through this process. 
Based on our internal testing, learning to integrate \name into an existing codebase and submitting it to our server for evaluation takes a participant less than one hour;
and an onward submission takes a single command line. 
Furthermore, \name can serve as a standalone tool for preparing the submission and providing basic efficiency metrics.

In providing abstractions around the evaluation interface, we limit assumptions made around the underlying system implementation and allow for the installation of user dependencies as needed.  This enables support for a diversity of backend frameworks and runtimes as the user is not constrained to a single deep learning framework or data format.  For example, \name allows users to use both research frameworks (e.g., eager execution PyTorch and TensorFlow 2.0) as well as specialized inference runtimes (e.g., ONNX Runtime, TVM, and TensorRT). The additional flexibility provided by this format allows \name to remain accessible to researchers familiar with a particular framework, while also enabling the exploration of different means of increasing overall \textit{end-to-end efficiency} of the machine learning system that is available in deployment settings. This design allows users to evaluate efficiency gains from improving different aspects of the overall system, such as those obtained from optimizing the model architectures or from utilizing faster software frameworks.

\name builds upon established software developed and maintained by AI2. 
These tools have been thoroughly tested by AI2 researchers and engineers, enhancing \name's robustness and ease of use. 
For example, empowered by Catwalk, \name supports a diverse set of NLP tasks, and allows \name to easily extend to many other tasks and research fields.\footnote{Catwalk provides a unified interface to a broad range of existing NLP tasks and models.  A list of tasks that are currently supported by \name can be found at \url{https://github.com/allenai/catwalk}.}

%% file: tables/scenarios.tex
\begin{table*}[ht]
\centering
\begin{tabular}{@{} l cccccccc @{}} 
\toprule
\bf Scenarios & \bf Acc. & \bf TP. & \bf Latency & \bf Mem. & \bf Energy \& CO$_2$ &  \bf BSZ &  \bf Online\\
\midrule 
\bf Fixed batching & \cmark & \cmark & \cmark & \cmark & \cmark  & User specified & \cmark \\

\bf Poisson batching & \xmark & \cmark & \cmark & \cmark & \cmark & Random & \cmark \\

\bf Single stream & \xmark & \xmark & \cmark & \cmark & \cmark & 1 & \cmark \\

\bf Offline & \xmark & \cmark & \xmark & \cmark & \cmark & User specified & \xmark \\
\midrule
\end{tabular}
\caption{Four evaluation scenarios and the metrics they focus on.
Acc.: accuracy, TP.: throughput, Mem.: memory.
In the three online scenarios, \name interfaces with the submitted model via standard input/output (\stdio), providing inputs and capturing outputs in real-time. 
Rearrangement of instance order is prohibited in these scenarios. 
In the offline scenario, the model is given immediate access to all evaluation instances via a file, enabling techniques such as sorting by lengths.
}
\vspace{-.5cm}
\end{table*}

%% file: text/experiments.tex
\section{Experiments}\label{sec:experiments}
We use \name to benchmark several established models for machine translation and text classification with the RAFT dataset \citep{alex2021raft}.
In the interest of space, we refer the readers to the appendices for the RAFT experiments.

\noindent \textbf{Machine Translation.}
Improving the efficiency of machine translation (MT) and text generation models has gained significant momentum.
A growing number of recent workshops and shared tasks have held dedicated efficiency tracks~\interalia{wngt2018,wngt2019,wngt2020,wmt21,wmt22}.
Aligned with this goal, we seek to contribute to this ongoing effort.
To this end, our initial experiments with \name focus on machine translation.


\noindent \textbf{Dataset and setting.}
We present results for \wmt~\citep{ws-2014-statistical}, a well-studied dataset that is selected as the testbed in the efficiency tracks of two recent WMT workshops~\citep{wmt21,wmt22}.
\name already supports many other MT and text generation datasets, and can be easily extended to more.
We focus on DE->EN translation here; additional results with EN->DE are available in the Appendices.

Balancing the inference wall clock time and accurately measuring the efficiency, we use different numbers of evaluating instances across the four scenarios. For \wmt:
\begin{itemize}[leftmargin=*,noitemsep,topsep=0pt,parsep=0pt,partopsep=0pt]
    \item {\bf Fixed batching} uses the full test set of 3,002 instances. It also measures the translation quality using SacreBLEU \citep{post-2018-call}.
    \item {\bf Poisson batching} randomly draws 4,000 instances (with replacement) from the test set. 
    \item In the {\bf single stream} scenario, 1,000 randomly selected test instances are used.
    \item Differently from others, the {\bf offline} scenario randomly selects 8,000 instances from the \emph{training} data.\footnote{In this scenario the models are granted immediate access to all instances and can sort them by length. If the instances \emph{were} drawn from the test set, this would result in the artifact that groups duplicates of the same instance in the same batch, which we aim to avoid.} We ensure that the selected instances have an average length matching that of the test set. 
\end{itemize}
Controlling for the random seed, all models are evaluated on the same set of instances in the same order, and identical batch sizes in the Poisson batching scenario. 
Preliminary experiments indicate that the models' efficiency performance remains consistent across multiple runs.
As such, we opt out of conducting multiple rounds of evaluation.
All models are evaluated on one RTX8000 GPU, and the inference batch sizes for the fixed batching and offline scenarios are tuned to the allowable maximum for the available GPU hardware.

\noindent \textbf{Models.}
We benchmark the following publicly-available models covering a wide range of sizes: 
\begin{itemize}[leftmargin=*,noitemsep,topsep=0pt,parsep=0pt,partopsep=0pt]
    \item {\bf MBART} \citep{tang-etal-2021-multilingual-mbart50}: a 610M-parameter-sized Transformer model for multilingual translation.
    It has two variants, many-to-one (MBART M2O) translates other languages into English, and many-to-many (M2M) can translate between multiple language pairs. We use the {\bf MBART50} variant, originally pre-trained on monolingual corpora in 25 languages, by fine-tuning on parallel corpora in across 50 languages for direct use as a translation engine.
    \item {\bf M2M100} \citep{m2m100}: Transformer-based multilingual models for many-to-many translation. We report on two sizes with 418M and 1.2B parameters respectively.
    The {\bf M2M100} model is trained using parallel corpora (e.g., WMT corpora described above) and mined bitext to enable translation between any two of 100 languages.
    \item {\bf OPUS} \citep{TiedemannThottingal:EAMT2020-OPUS}: a bilingual Transformer model with 74M parameters for DE->EN translation. The model is trained on OPUS bitext corpora~\citep{tiedemann2012parallel}. 
    \item {\bf WMT19-Meta} \citep{ng2019facebook-wmt19-de-en}: a DE->EN Transformer model with 314M parameters. 
    
    This system won the WMT19 task on German to English news translation \citep{wmt19}.
    \item {\bf WMT21-Meta} \citep{tran2021facebook-wmt21-dense-de-en}: a M2O Transformer model with 4.7B parameters. 
    Unlike {\bf WMT19-Meta}, this model is multilingual and trained on data from all languages for the WMT 2021 shared task.
    Training data is a mixture of parallel corpora, monolingual corpora and mined bitext. This multilingual system ranked high in several WMT21 news translation tasks \citep{wmt21}. We refer to  \citet{tran2021facebook-wmt21-dense-de-en} for complete details.
\end{itemize}
We evaluate using PyTorch with both full precision (FP32) and half precision (FP16), to study the effect of quantization. In our preliminary experiments, we found that employing more aggressive quantization techniques such as 8-bit and 4-bit quantization using naive methods led to severely compromised translation quality, with the BLEU score dropping to around 1, effectively resulting in a failed translation.
All models' implementation and checkpoints are available on Hugging Face.

\begin{figure*}[t!]
\centering
\begin{subfigure}[t]{0.48\textwidth}
\includegraphics[width=\textwidth, trim={9.4cm 3.5cm 9.5cm 3.5cm},clip]{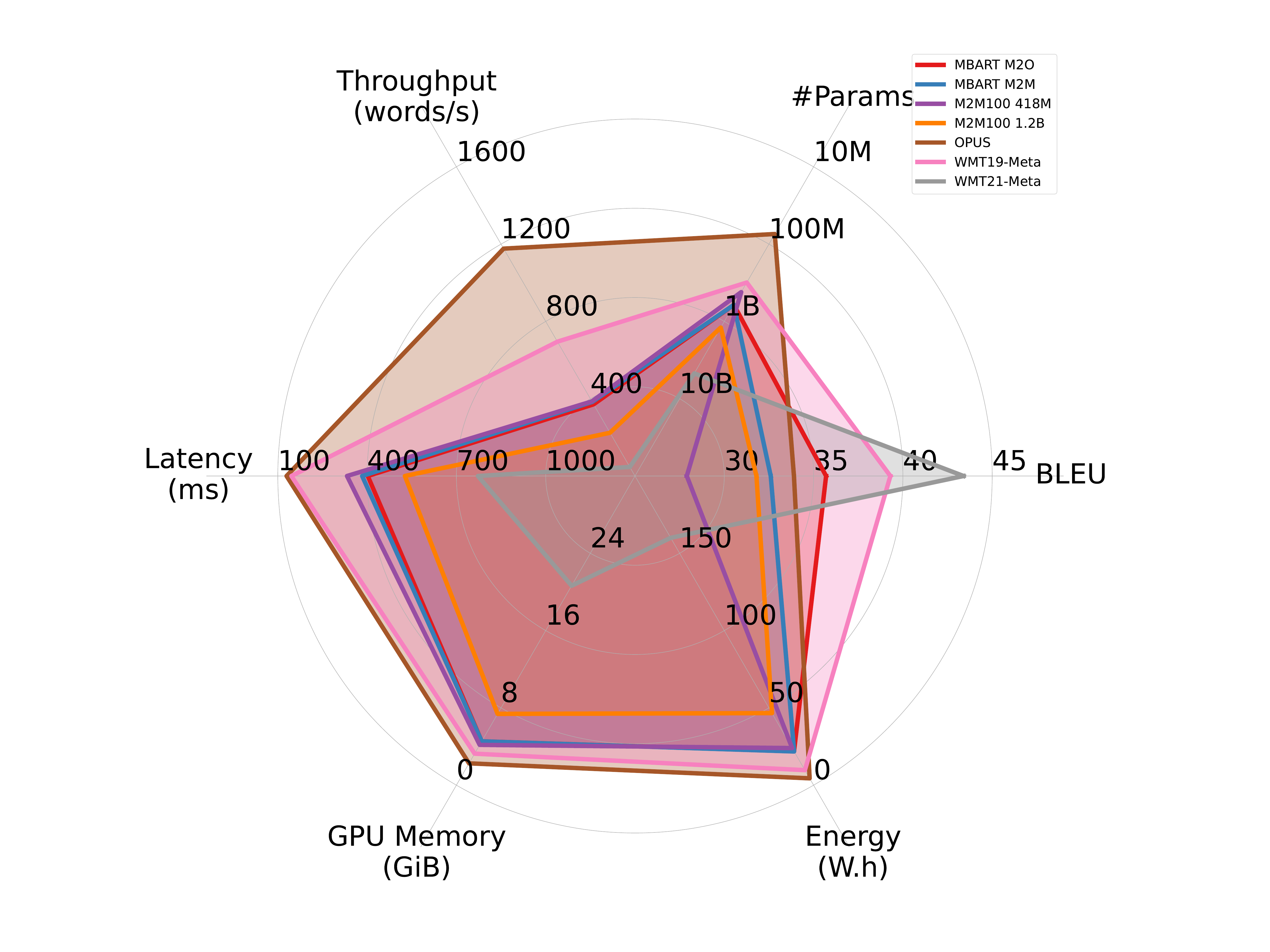}
\caption{FP32. }
\label{fig:wmt14-de-en-fp32}
\end{subfigure}
\hfill
\begin{subfigure}[t]{0.48\textwidth}
\includegraphics[width=\textwidth, trim={9.4cm 3.5cm 9.5cm 3.5cm},clip]{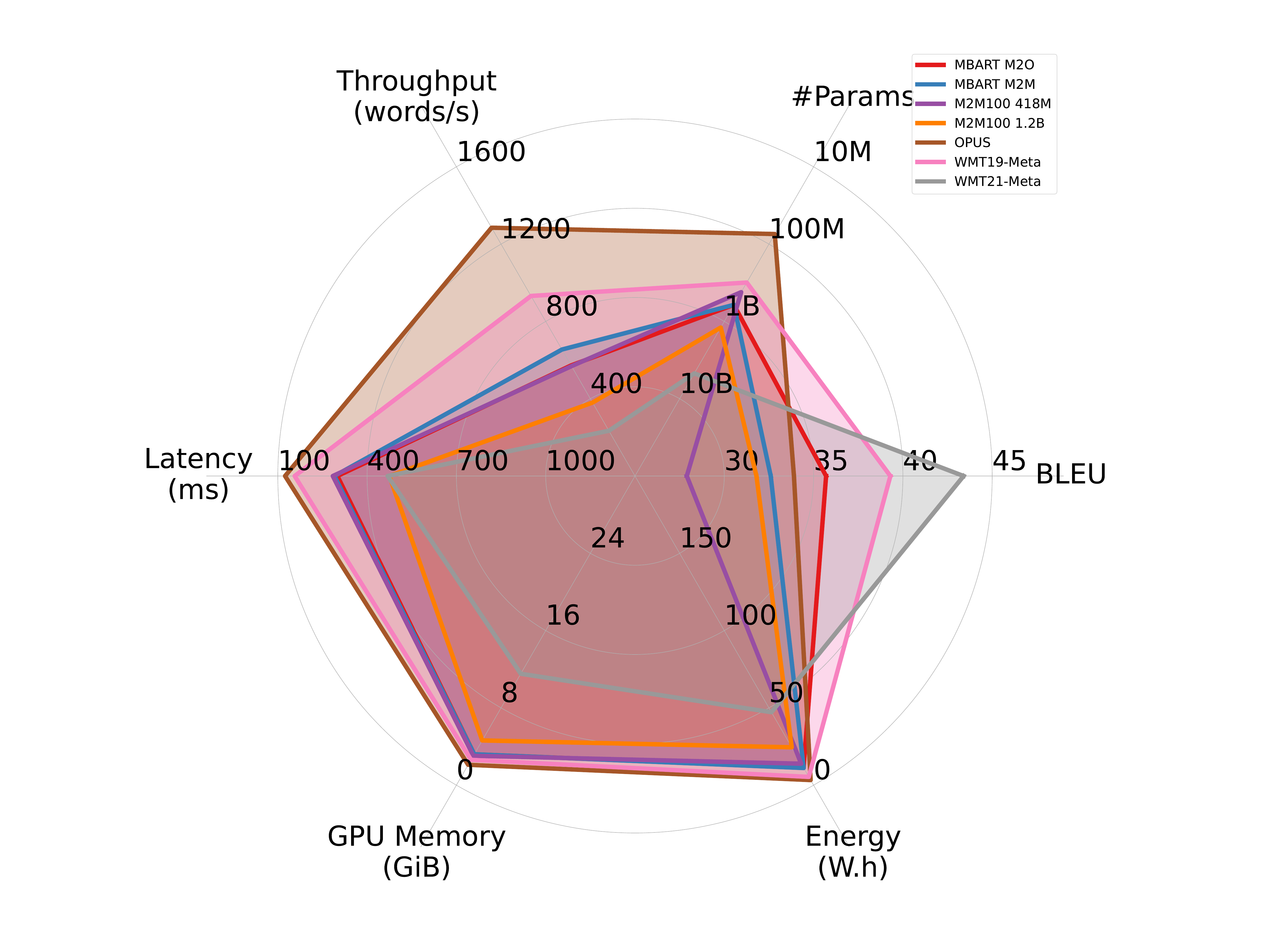}
\caption{FP16 }
\label{fig:wmt14-de-en-fp16}
\end{subfigure}
~
\begin{subfigure}[t]{0.48\textwidth}
\includegraphics[width=\textwidth, trim={9.4cm 3.5cm 9.5cm 3.5cm},clip]{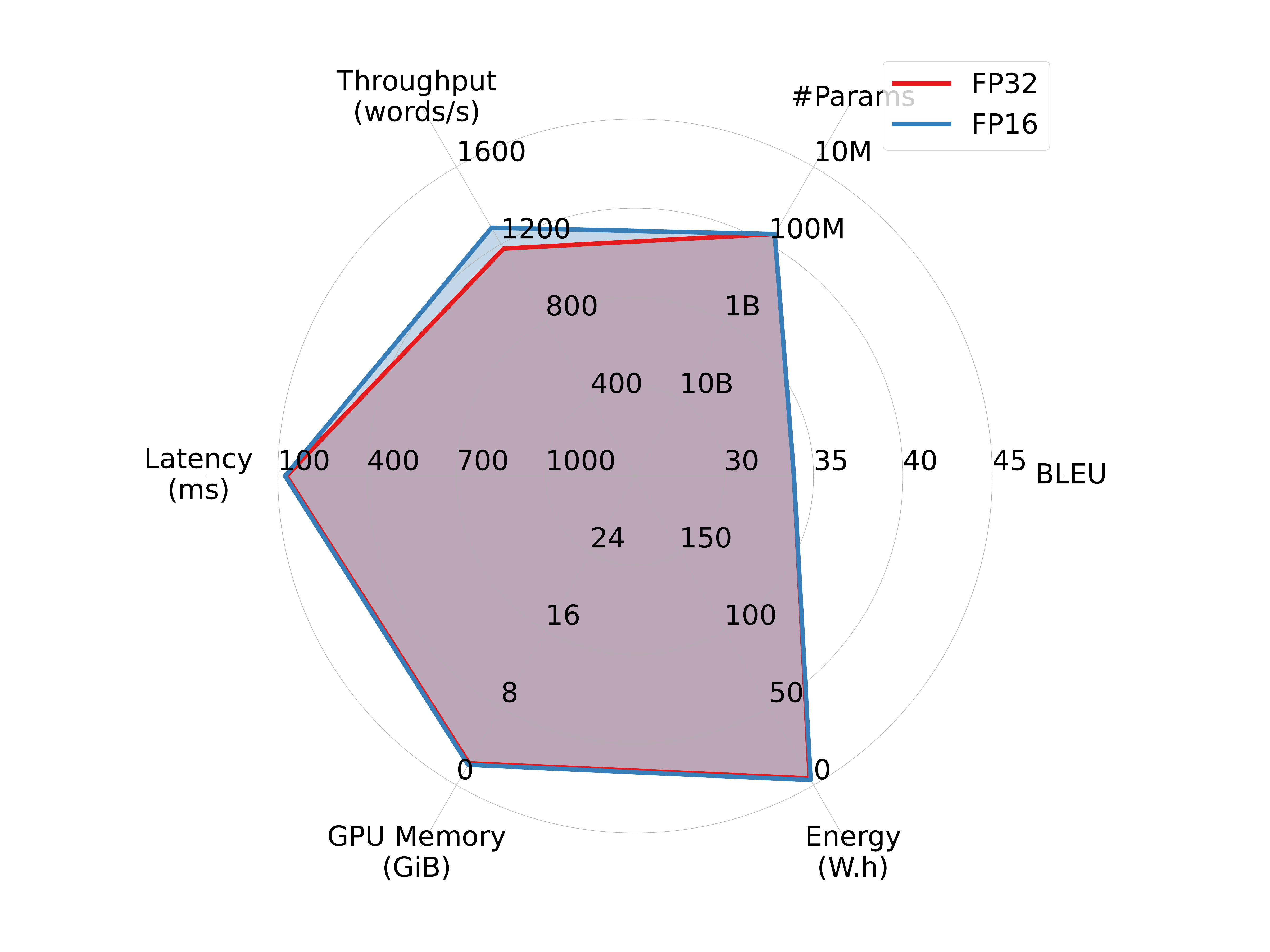}
\caption{OPUS, FP32 vs. FP16. }
\label{fig:wmt14-de-en-opus}
\end{subfigure}
\hfill
\begin{subfigure}[t]{0.48\textwidth}
\includegraphics[width=\textwidth, trim={9.4cm 3.5cm 9.5cm 3.5cm},clip]{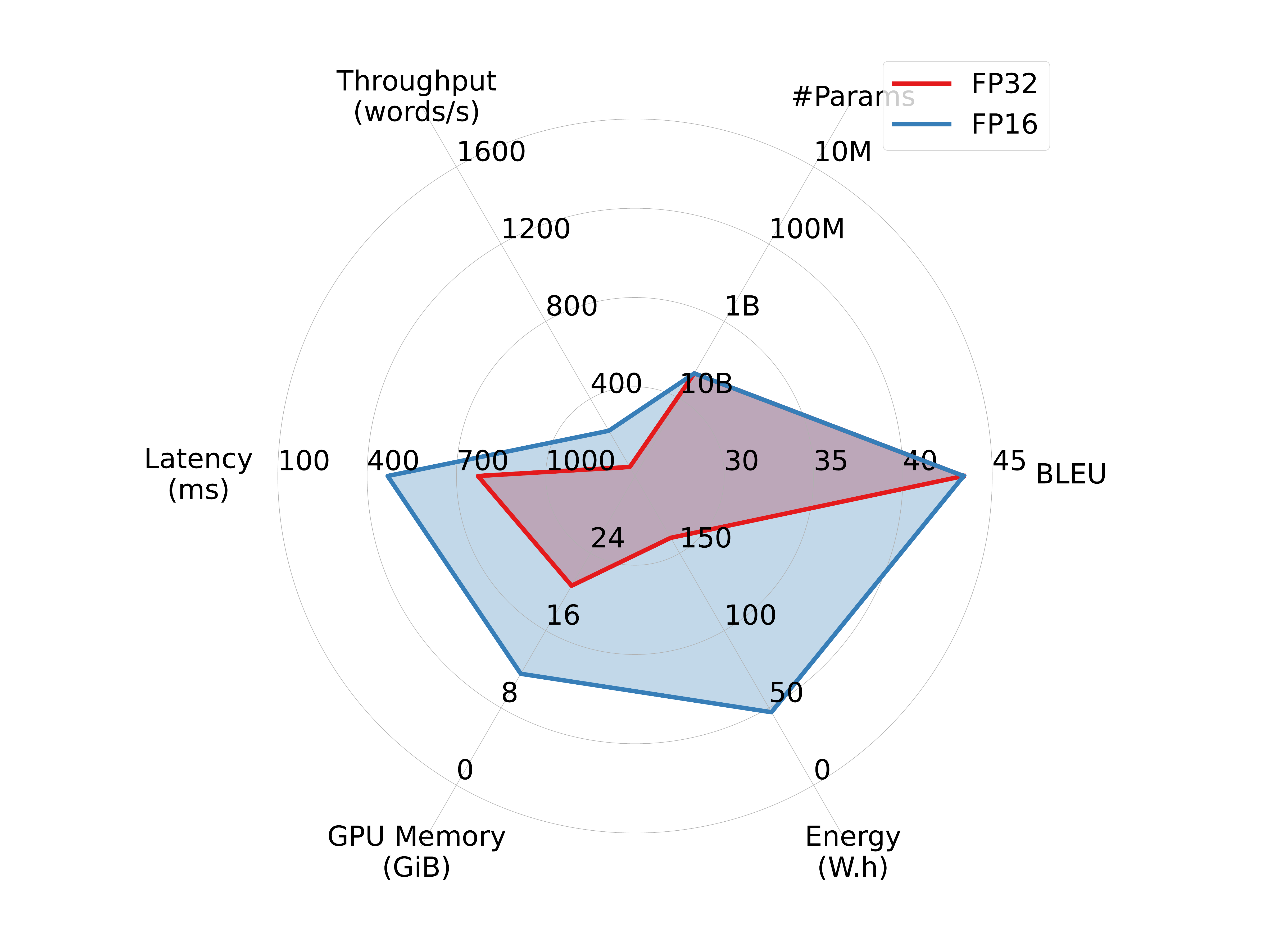}
\caption{WMT21-Meta, FP32 vs. FP16. }
\label{fig:wmt14-de-en-wmt21-meta}
\end{subfigure}
\caption{Performance of various models on the \wmt, represented in terms of BLEU scores and a range of efficiency metrics. To more accurately reflect real-world applications, the figures include throughput metrics from the offline scenario, latency and GPU memory metrics from the single stream scenario, and energy metrics from the fixed batching scenario. For all metrics, {\bf outer rings indicate better performance}. \#Params is presented on a logarithmic scale.}
\label{fig:radar}
\vspace{-.5cm}
\end{figure*}

\noindent\textbf{Results.}
Figure~\ref{fig:radar} summarizes the efficiency performance of different models in on the \wmt dataset, along with their translation quality. 
Overall, models trained for English translation demonstrated better trade-offs between translation quality and efficiency. 
Notably, OPUS outperforms the much larger MBART M2M and M2M100 models in both accuracy and all aspects of efficiency, and is the most efficient model among all.
Although WMT21-Meta, the largest model considered, provides the highest BLEU score, it takes a substantial hit in efficiency. 

Interestingly, despite being more than four times larger,
WMT19-Meta achieves efficiency performance comparable to OPUS in latency, memory overhead, and energy consumption, and significantly outperforms it in terms of BLEU.
However, it falls short of OPUS in throughput.
This observation confirms that relying on a single efficiency metric risks oversimplifying the complex performance landscape of efficiency in practical applications.

With ONNX, the models achieve over 20\% improvements in latency and throughput in the single-stream scenario, accompanied by a significant reduction in memory and energy overhead.
However, less efficiency improvement is observed in other scenarios with larger batch sizes. 

\noindent \textbf{Larger models benefit more from FP16 quantization.}
By comparing Figures~\ref{fig:wmt14-de-en-fp32} and \ref{fig:wmt14-de-en-fp16}, we observe that FP16 quantization improves all models' efficiency performance (except \#Params.), particularly memory overhead. 
Larger models appear to benefit more from quantization. 
As shown in Figures~\ref{fig:wmt14-de-en-opus} and \ref{fig:wmt14-de-en-wmt21-meta}, while OPUS experiences minimal efficiency gains from quantization apart from increased throughput, WMT21-Meta's efficiency dramatically improves with FP16 quantization, nearly doubling throughput and reducing latency, memory overhead, and energy consumption by half or more. 
These results highlight the promise of advancing quantization techniques for larger models in order to improve the trade-off between accuracy and efficiency.

\begin{figure*}[t!]
\centering
\begin{subfigure}[t]{0.49\textwidth}
\includegraphics[width=\textwidth, trim={1cm 3cm 5cm 2.5cm},clip]{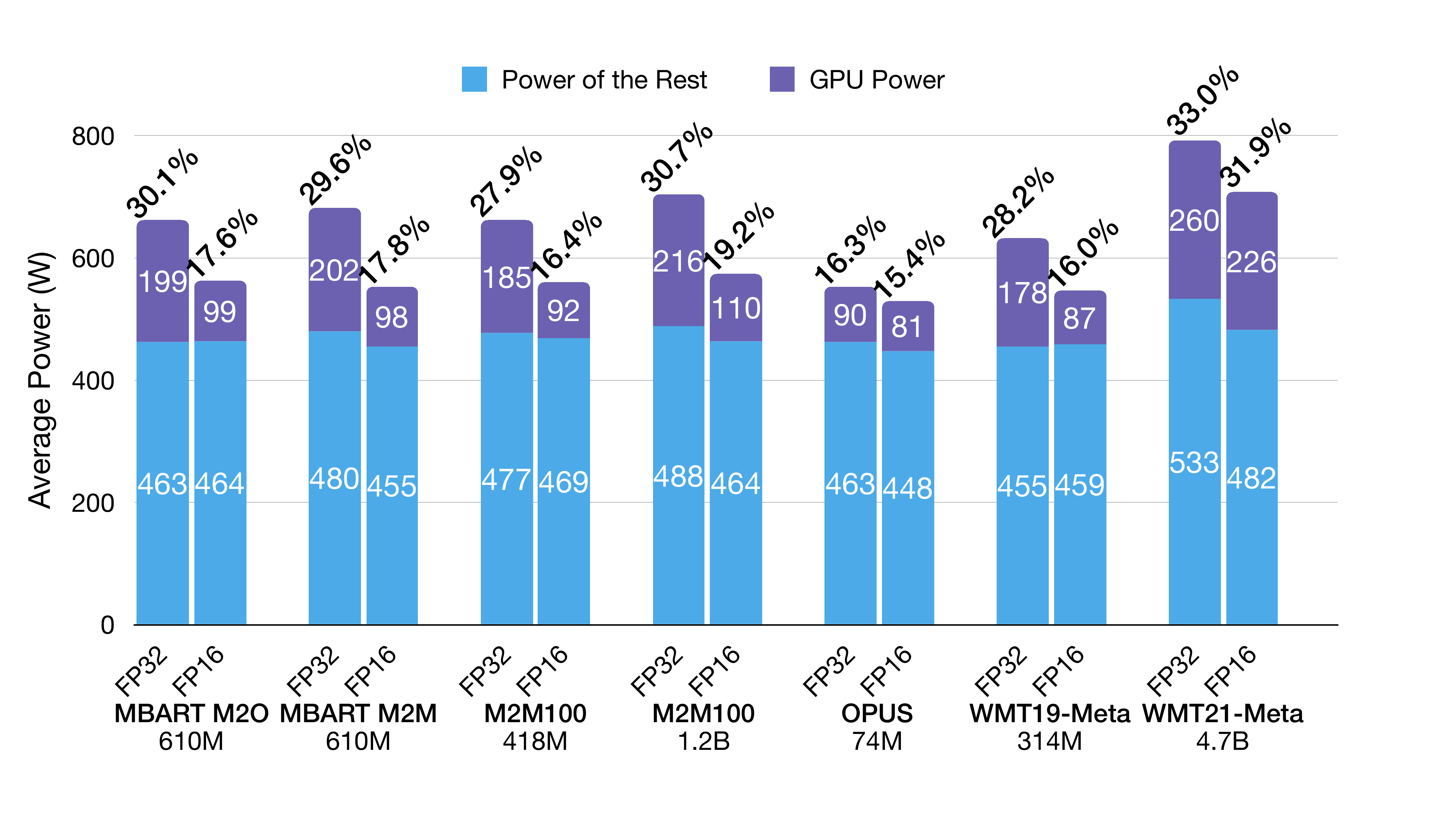}
\caption{Single stream.}
\label{fig:single-stream}
\end{subfigure}
\hfill
\begin{subfigure}[t]{0.49\textwidth}
\includegraphics[width=\textwidth, trim={1cm 3cm 5cm 2.5cm},clip]{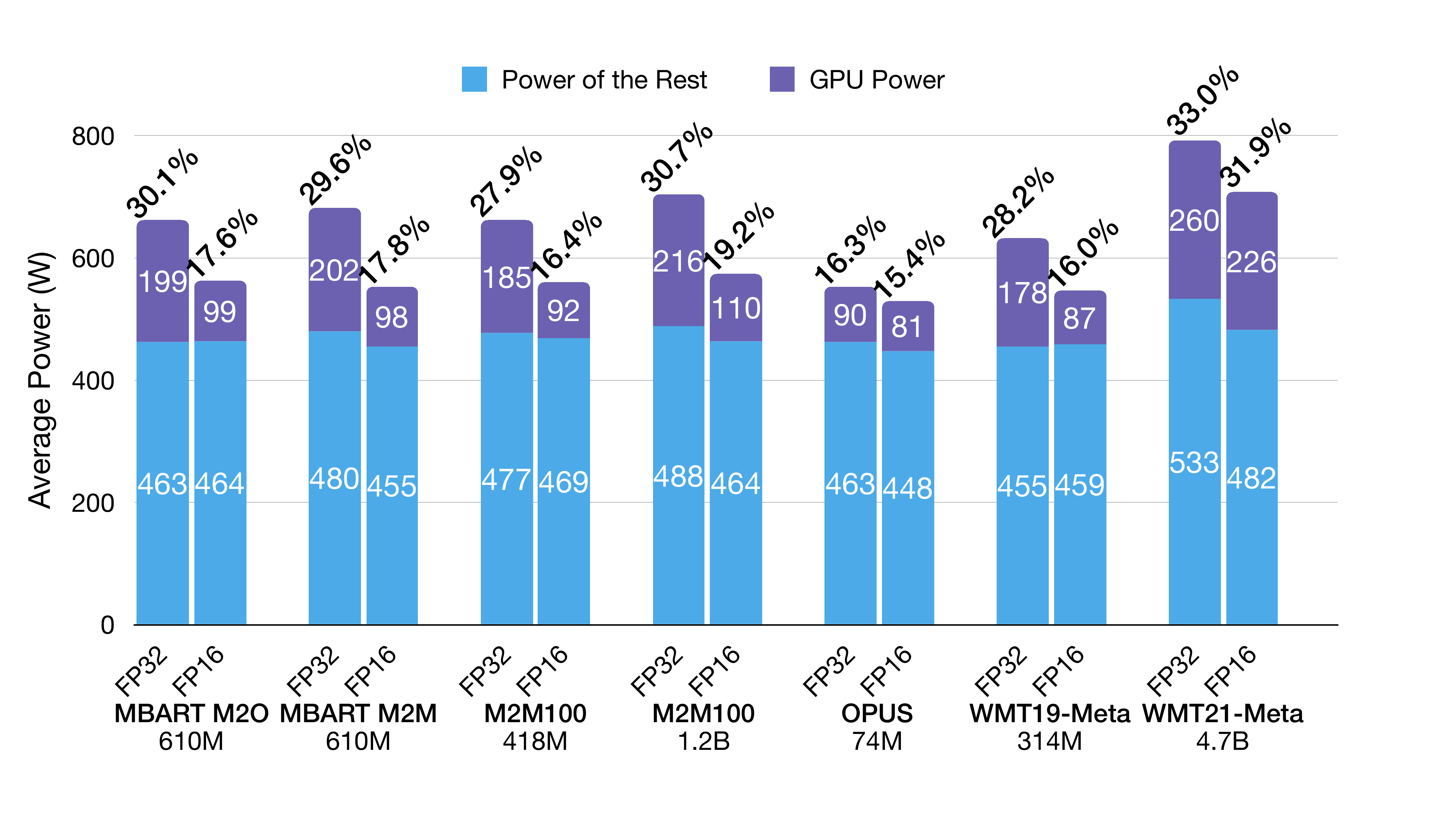}
\caption{Offline.}
\label{fig:offline}
\end{subfigure}
\caption{
Power consumption in Watts across different model inference runs in the single stream (\ref{fig:single-stream}) and offline (\ref{fig:offline}) scenarios. Purple bars indicate the power consumed by the GPU, while the light blue bars represent the power consumption of all other system components, excluding the GPU. The white numbers denote the absolute power consumption values in Watts, while the percentage numbers atop the bars provide the proportion of power consumption that is accounted for by the GPU. 
}
\label{fig:power}
\vspace{-.5cm}
\end{figure*}

\noindent \textbf{In single-GPU inference, the GPU accounts for only a minor portion of the energy consumption.}
This is demonstrated by Figure~\ref{fig:power}. 
This experiment uses a single RTX8000 GPU with a maximum power of 260W. 
We note that the GPU rarely operates at full power, implying that GPU hours, a metric commonly used to gauge training computational overhead~\citep{henderson2020towards,kasai2021t2r}, is unsuitable for estimating inference GPU energy. 
Even during the most GPU-intensive runs by the WMT21-Meta model, where it does operate at full capacity, the GPU only accounts for one third of the total machine power.
This observation diverges from previous findings on \emph{training}, where GPUs are estimated to constitute around 70\% of the energy usage~\citep{dodge2022measuring}. 
We attribute the difference to the increased memory and disk IO demands during inference, coupled with 
lower GPU utilization and increased idling time due to smaller compute kernels during inference
This disparity suggests that efficiency conclusions drawn from training need careful examination when applied to inference.  
Interestingly, we observe a correlation between higher GPU power and higher power utilization by other components. 
We conjecture that this is at least partially due to the increased fan activity needed for cooling.

%% file: text/related.tex
\section{Related Work}
There is growing interest in putting efficiency in NLP benchmarks.
Dynabench~\citep{kiela-etal-2021-dynabench} and Dynaboard~\citep{maDynaboardEvaluationAsAServicePlatform2021} concentrate on dynamic dataset creation and model assessment, incorporating efficiency metrics such as throughput and memory, alongside fairness and robustness
HELM~\citep{liang2022holistic} evaluates language models with seven metrics including efficiency. Though training efficiency in HELM covers energy, carbon, and wallclock time, the inference efficiency in this benchmark only measures inference runtime, and the energy and carbon footprint are only roughly estimated.
HULK~\citep{zhou-etal-2021-hulk} evaluates energy efficiency as a proxy of time and cost, while \name evaluates multiple different efficiency metrics in a realistic way. 
Long-Range Arena~\citep{tay2020long} builds a set of synthesized tasks to evaluate the long-range capabilities of NLP models in terms of generalization and computational efficiency including speed and memory footprint.  
Another line of work has studied application- or task-specific efficiency such as trade-offs between accuracy and energy consumption for long context NLP models~\citep{ang-etal-2022-characterizing}, inference energy competition for models on SuperGLUE~\citep{wang-wolf-2020-overview} or storage efficiency for open domain question answering~\citep{pmlr-v133-min21a}.
Most related to \name, MLPerf targets inference efficiency across various real-world scenarios~\citep{2020-mlperf-reddi, banburymlperf, mattson2020mlperf}.
While MLPerf aims to stimulate building more efficient hardware platforms, \name incentivizes algorithmic innovations, controlling the hardware.
Hosted on an in-house machine, \name can accurately measure inference energy consumption, which was impossible for previous benchmark efforts.






%% file: text/conclusion.tex
\section{Conclusions}


We present \name, a benchmark for holistic and realistic evaluation of inference efficiency. 
\name targets multiple aspects of efficiency including latency, throughput, memory overhead, number of parameters, and energy consumption, on a strictly-controlled
hardware platform. 
Integrating evaluation with \name is seamless and can drastically reduce the workload to make fair and reproducible efficiency comparisons. 
\name offers both testing in real-world application scenarios and a standardized platform for comparison between any two submissions. 
We establish this tool for NLP models but offer flexible extensions to additional tasks and scenarios. 
We envision \name to provide a new lens on testing algorithmic innovations by lowering the barrier to entry for evaluating efficiency and characterizing
environmental impact of future models.

%% file: text/checklist.tex
\section*{Checklist}


\begin{enumerate}

\item For all authors...
\begin{enumerate}
  \item Do the main claims made in the abstract and introduction accurately reflect the paper's contributions and scope?
    \answerYes{}
  \item Did you describe the limitations of your work?
    \answerYes{
    As mentioned in Section ~\ref{sec:controlling}, our current server for \name houses two Nvidia RTX 8000 GPUs. We plan to support other hardware platforms such as the edge Nvidia Jetson TX2 in the near future.
    Our current GPUs are based on the previous generation of Turing microarchitecture, which might not fully utilize the state-of-the-art GPU technology and CUDA software improvements.
    To address this, we have plans to upgrade our server with more advanced GPUs in the near future.

    Our requirements for submission of code and checkpoints naturally prohibit the evaluation of large language models that are not publicly available, or any large model that our hardware is not capable of running. In addition to the plan to upgrade our hardware, the insights learned from \name could help better quantify hardware's impact on efficiency and possibly extrapolate the findings to the models we are currently unable to evaluate.

    Finally, while \name focuses on evaluating inference efficiency, we acknowledge the challenges and complexity of properly evaluating training efficiency.
    Our hope is that the insights and methodologies developed through \name can also contribute to improved tools and strategies for evaluating and comparing the training efficiency of large models in the future.
    }
  \item Did you discuss any potential negative societal impacts of your work?
    \answerYes{The goal of this work is to mitigate the negative societal impacts of ML due to the increasing computational demands of ML models, by providing a better platform for benchmarking model efficiency. We do not anticipate any direct negative societal impacts. There may be potential indirect impacts if our platform facilitates drastic improvements in ML model efficiency, leading to e.g. (a) increased overall emissions due to increased ease of use/access (i.e. Jevons paradox) or (b) increased access to ML models by bad actors, who would have otherwise been limited by computational resources. In the case of (b), we hope that any increased access facilitated by our work is equally applicable to good actors, balancing the effect.}
  \item Have you read the ethics review guidelines and ensured that your paper conforms to them?
    \answerYes{}
\end{enumerate}

\item If you are including theoretical results...
\begin{enumerate}
  \item Did you state the full set of assumptions of all theoretical results?
    \answerNA{}
	\item Did you include complete proofs of all theoretical results?
    \answerNA{}
\end{enumerate}

\item If you ran experiments (e.g. for benchmarks)...
\begin{enumerate}
  \item Did you include the code, data, and instructions needed to reproduce the main experimental results (either in the supplemental material or as a URL)?
    \answerYes{The URL to code is provided in Section~\ref{sec:intro}.}
  \item Did you specify all the training details (e.g., data splits, hyperparameters, how they were chosen)?
    \answerYes{We use the standard data splits for evaluation with the WMT14 DE-EN dataset. We do not perform model training in this work.}
	\item Did you report error bars (e.g., with respect to the random seed after running experiments multiple times)?
    \answerNo{We do not perform training, so there are no averages over random seeds. As discussed in Section~\ref{sec:experiments}, we found in preliminary experiments that there was little variation in efficiency measures across runs, so we run each evaluation setting only once.}
	\item Did you include the total amount of compute and the type of resources used (e.g., type of GPUs, internal cluster, or cloud provider)?
    \answerYes{Reported or computable from results reported in Section~\ref{sec:experiments}.}
\end{enumerate}

\item If you are using existing assets (e.g., code, data, models) or curating/releasing new assets...
\begin{enumerate}
  \item If your work uses existing assets, did you cite the creators?
    \answerYes{See Section~\ref{sec:experiments}.}
  \item Did you mention the license of the assets?
    \answerNo{The license of the WMT14 DE-EN dataset is not listed on the dataset website. However, data is sourced from the EuroParl corpus with no listed copyright restrictions.\footnote{For more information see the WMT-14 Task \url{https://www.statmt.org/wmt14/translation-task.html} and the EuroParl Corpus \url{https://www.statmt.org/europarl/}}}
  \item Did you include any new assets either in the supplemental material or as a URL?
    \answerNo{}
  \item Did you discuss whether and how consent was obtained from people whose data you're using/curating?
    \answerNo{We use WMT14 DE-EN, a popular long-standing corpus of parallel translation data. The curators of this dataset do not report this information.}
  \item Did you discuss whether the data you are using/curating contains personally identifiable information or offensive content?
    \answerNo{We use WMT14 DE-EN, a popular long-standing corpus of parallel translation data. The curators of this dataset do not report this information.}
\end{enumerate}

\item If you used crowdsourcing or conducted research with human subjects...
\begin{enumerate}
  \item Did you include the full text of instructions given to participants and screenshots, if applicable?
    \answerNA{}
  \item Did you describe any potential participant risks, with links to Institutional Review Board (IRB) approvals, if applicable?
    \answerNA{}
  \item Did you include the estimated hourly wage paid to participants and the total amount spent on participant compensation?
    \answerNA{}
\end{enumerate}

\end{enumerate}

%% file: text/appendix.tex
\begin{appendices}

\section{Text Classification with RAFT}
RAFT is a collection of 11 datasets that focus on few-shot text classification in real-world settings.
Here we focus on the ADE Corpus V2 (ADE) portion, aiming to classify sentences derived from medical reports as related or unrelated to adverse drug effects.
Several baseline models, provided by the authors, were evaluated for efficiency, including:
\begin{itemize}[leftmargin=*,noitemsep,topsep=0pt,parsep=0pt,partopsep=0pt]
    \item {\bf AdaBoost}~\citep{adaboost}: a strong non-neural classifier based on decision trees. 
    \item {\bf BART Zero Shot MNLI}: BART \citep{lewis-etal-2020-bart} finetuned on the MNLI dataset \citep{williams-etal-2018-broad}. It is used as a zero-shot classifier.
    \item {\bf GPT-2}~\citep{radford2018language}: used as a few-shot classifier with 25 in-context training demonstrations and task-specific instructions.
\end{itemize}
The implementation for all models is attributed to \citet{alex2021raft}.\footnote{\url{https://github.com/oughtinc/raft-baselines}}
At the time of writing, RAFT has not released the gold labels of the test split, and therefore we report the F1 performance by \citet{alex2021raft}.

\begin{wrapfigure}[12]{r}{0.4\textwidth}
\vspace{-1cm}
\includegraphics[width=0.4\textwidth, trim={9.5cm 3.5cm 9.5cm 3.5cm},clip]{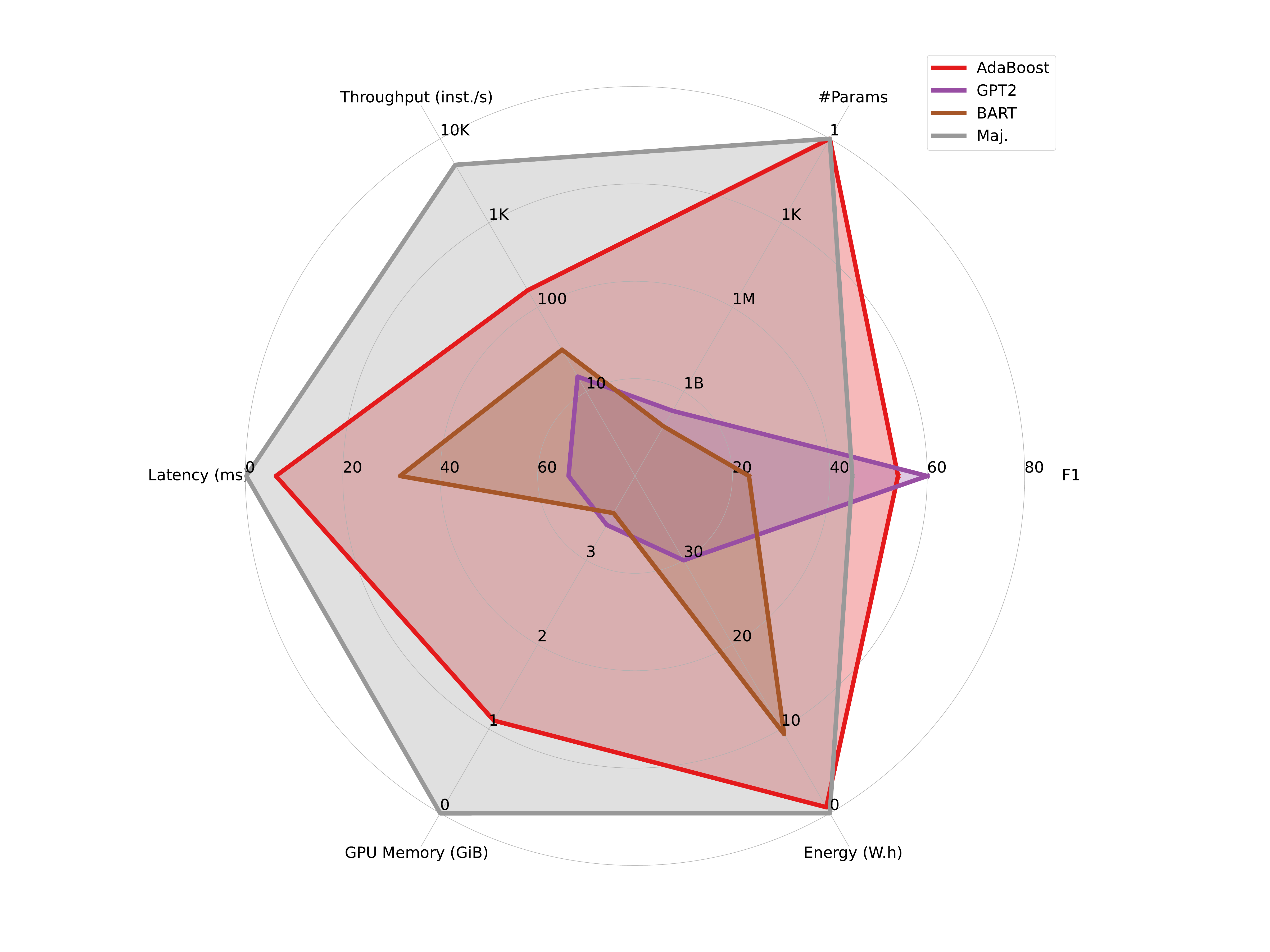}
\caption{Models' efficiency and accuracy performance on RAFT test data. F1 numbers are due to \citet{alex2021raft}.}
\label{fig:raft}
\end{wrapfigure}
\noindent \textbf{Results.}
Figure~\ref{fig:raft} provides a comparison of the above models with a majority-class baseline (Maj.). 
AdaBoost, a non-neural model, emerges as a strong competitor in terms of the accuracy-efficiency trade-off. 
Interestingly, GPT-2, despite having fewer parameters, lags behind BART in terms of throughput, latency, and energy consumption. 
We believe that this could be due to the in-context few-shot examples, which lead to significantly longer inputs for GPT-2 compared to BART. 
Nonetheless, GPT-2 manages to achieve the highest F1 score in this experiment.

\section{Additional Experiments with Machine Translation}

\begin{figure*}[t]
\centering
\begin{subfigure}[t]{0.48\textwidth}
\includegraphics[width=\textwidth, trim={9.4cm 3.5cm 9.5cm 3.5cm},clip]{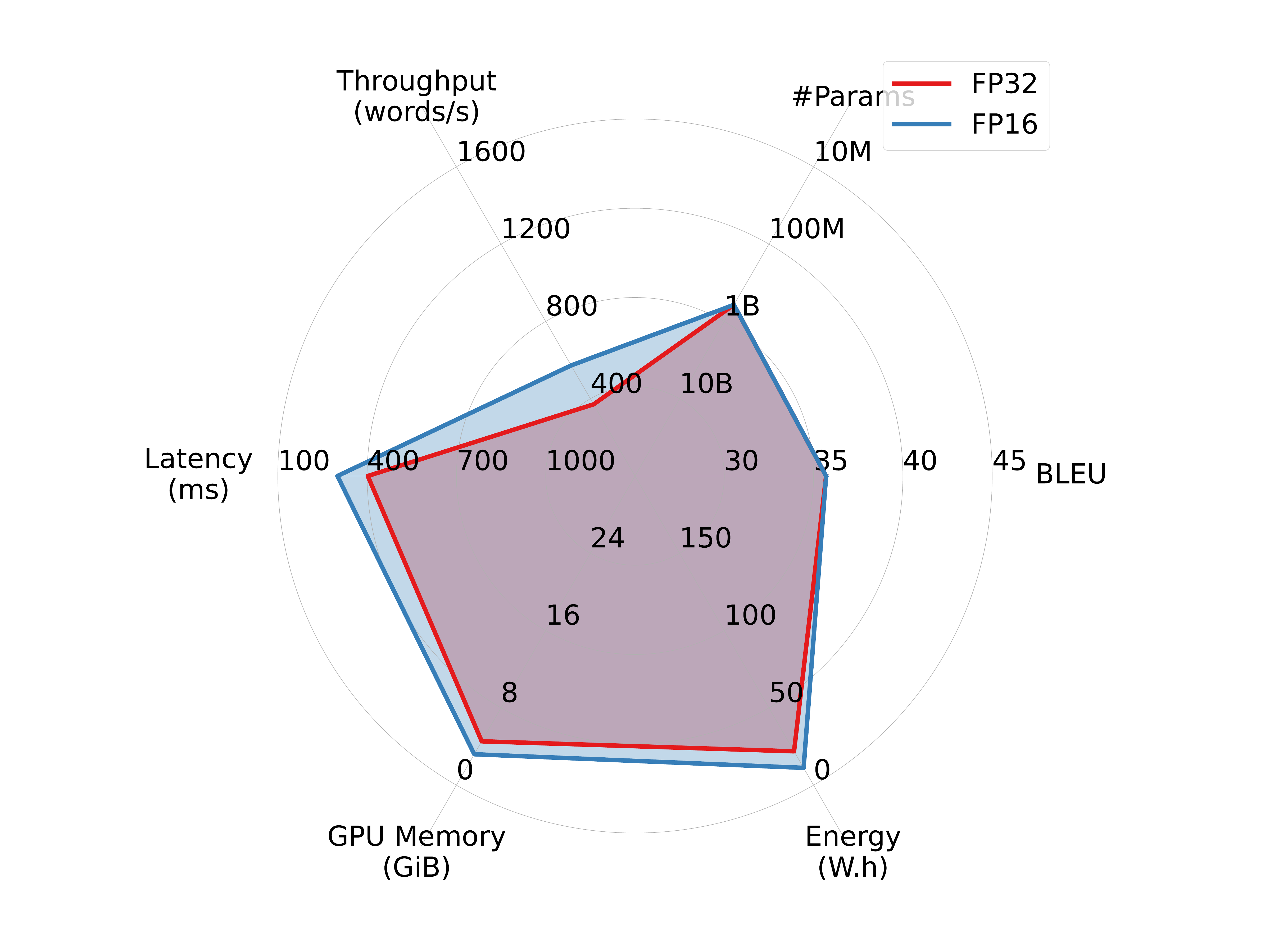}
\caption{MBART M2O. }
\end{subfigure}
\hfill
\begin{subfigure}[t]{0.48\textwidth}
\includegraphics[width=\textwidth, trim={9.4cm 3.5cm 9.5cm 3.5cm},clip]{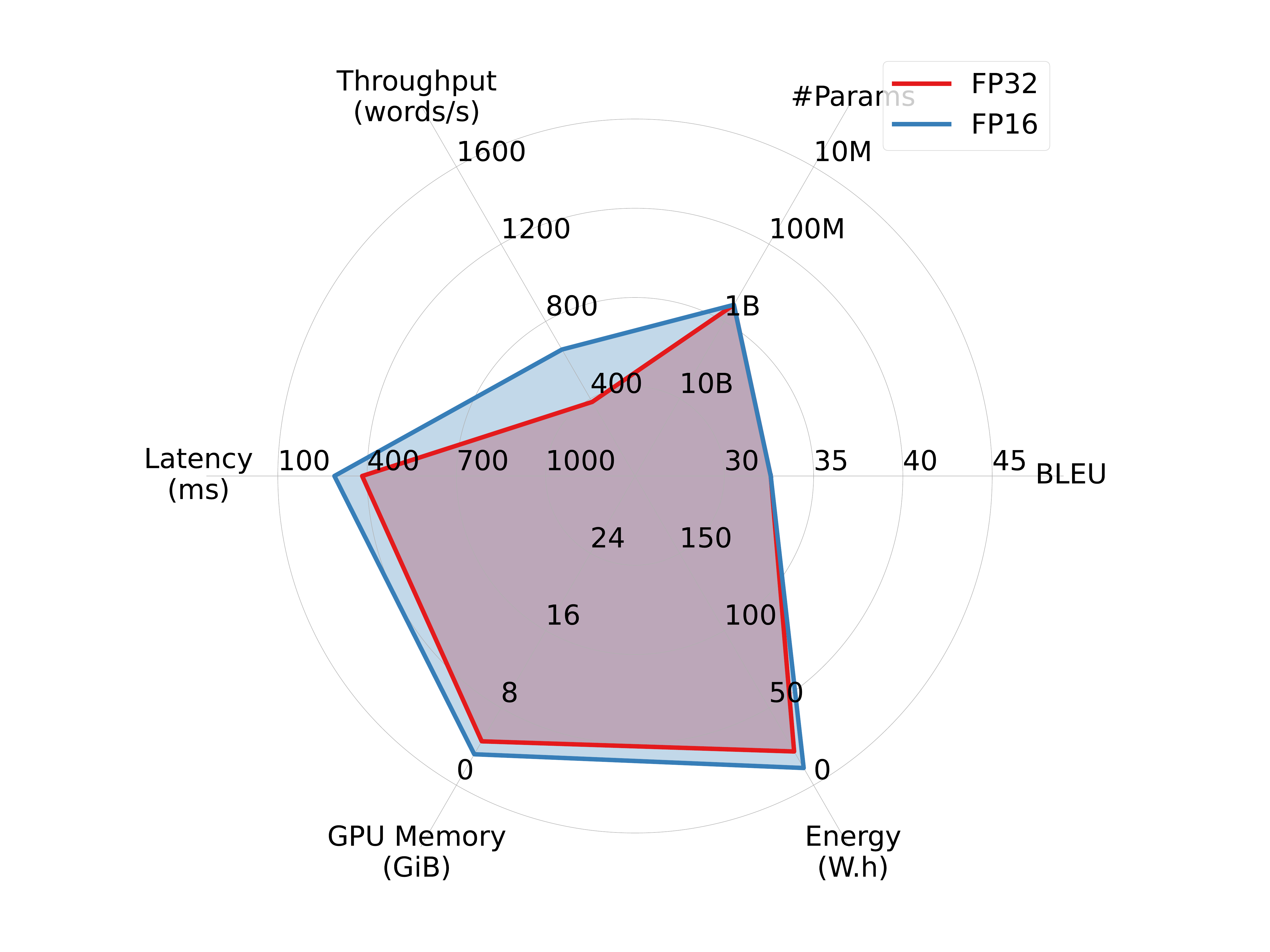}
\caption{MBART M2M. }
\end{subfigure}
~
\begin{subfigure}[t]{0.48\textwidth}
\includegraphics[width=\textwidth, trim={9.4cm 3.5cm 9.5cm 3.5cm},clip]{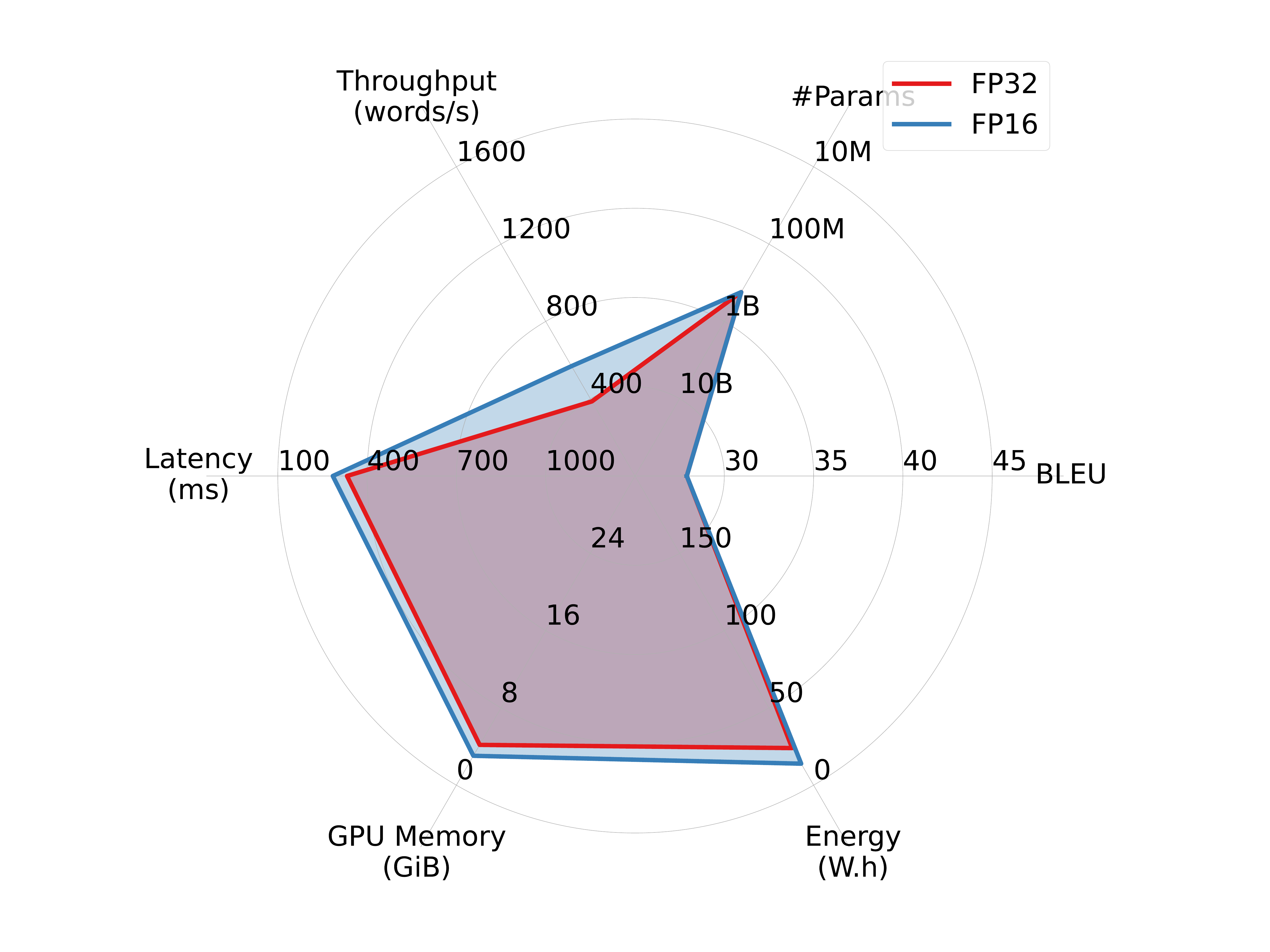}
\caption{M2M100 418M. }
\end{subfigure}
\hfill
\begin{subfigure}[t]{0.48\textwidth}
\includegraphics[width=\textwidth, trim={9.4cm 3.5cm 9.5cm 3.5cm},clip]{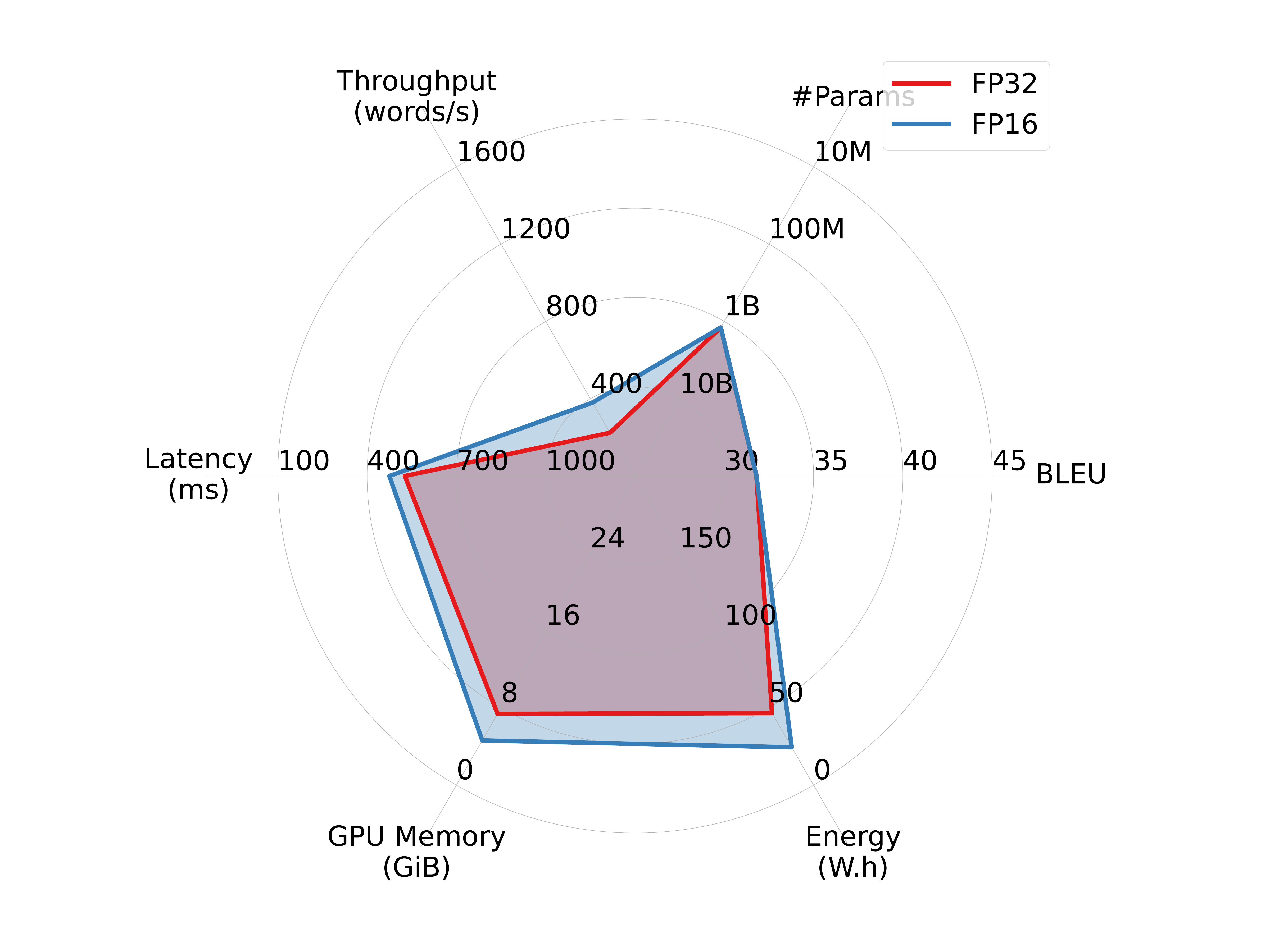}
\caption{M2M100 1.2B. }
\end{subfigure}
~
\begin{subfigure}[t]{0.48\textwidth}
\includegraphics[width=\textwidth, trim={9.4cm 3.5cm 9.5cm 3.5cm},clip]{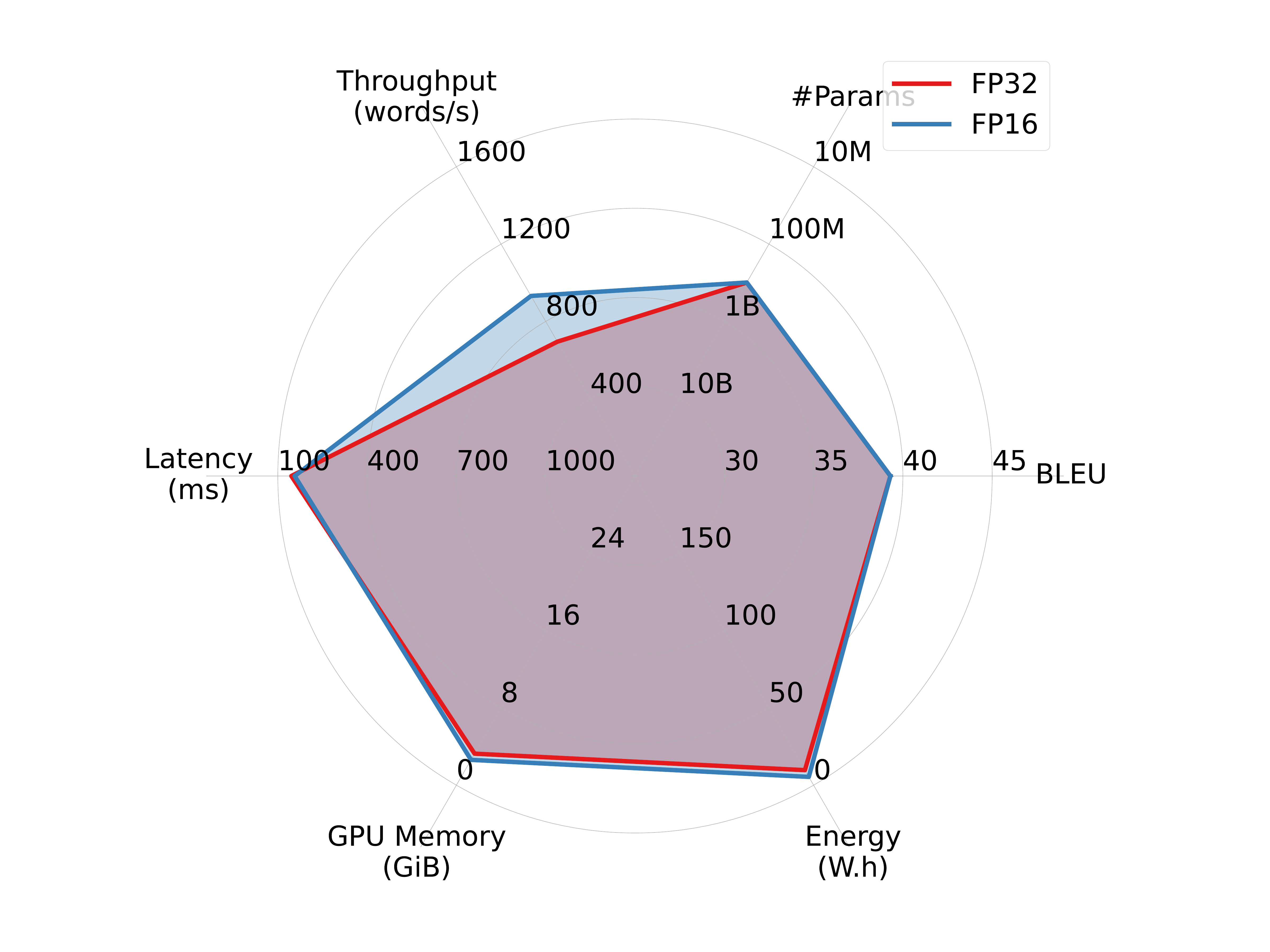}
\caption{WMT19 Meta. }
\end{subfigure}
\hfill
\caption{Additional results of various models on the \wmt using FP32 (red) and FP16 (blue). Similarly to Figure~\ref{fig:radar}, the throughput metrics are from the offline scenario, latency and GPU memory metrics from the single stream scenario, and energy metrics from the fixed batching scenario.}
\label{fig:additional_fp32_fp16}
\end{figure*}

All models' implementation and checkpoints are available on Hugging Face, with the following identifiers:
\begin{itemize}
    \item  MBART50: \texttt{facebook/mbart-large-50-many-to-\{many, one\}-mmt};
    \item M2M100: \texttt{facebook/m2m100\_\{418M, 1.2B\}};
    \item OPUS: \texttt{Helsinki-NLP/opus-mt-de-en};
    \item WMT19-Meta: \texttt{facebook/wmt19-de-en};
    \item WMT21-Meta: \texttt{facebook/wmt21-dense-24-wide-en-x}.
\end{itemize}

\noindent\textbf{Additional FP32 vs. FP16 comparisons.}
Figure~\ref{fig:additional_fp32_fp16} provides an additional set of comparisons between FP32 and FP16 across various models on \wmt, complementing the results presented in Section~\ref{sec:experiments}. The general trends mirror those observed earlier, with larger models benefiting more in terms of efficiency from quantization compared to smaller ones.

\noindent\textbf{ONNX improves throughput, latency, and energy overhead, at the cost of increased GPU memory overhead.}
\name makes little assumptions on the models' implementation and backend runtime,
and allows users to use both eager-execution research frameworks like PyTorch as well as specialized inference runtimes like Open Neural Network Exchange (ONNX).
Here we study ONNX's impact on the model's efficiency. 

ONNX is a cross-platform static runtime that uses pre-compiled computational graphs.
It allows for aggressive, global ahead-of-time compiler optimizations,
and can bring substantial latency and throughput improvements in inference settings with small batch size.
The readers are referred to \url{https://onnx.ai/} for more details.
As of now, ONNX supports conversion from models implemented with PyTorch, Tensorflow, and JAX, enabling us to make direct comparisons between PyTorch implementation and ONNX in our machine translation experiments with \wmt.

\begin{figure*}[t]
\centering
\begin{subfigure}[t]{0.48\textwidth}
\includegraphics[width=\textwidth, trim={9.4cm 3.5cm 9.5cm 3.5cm},clip]{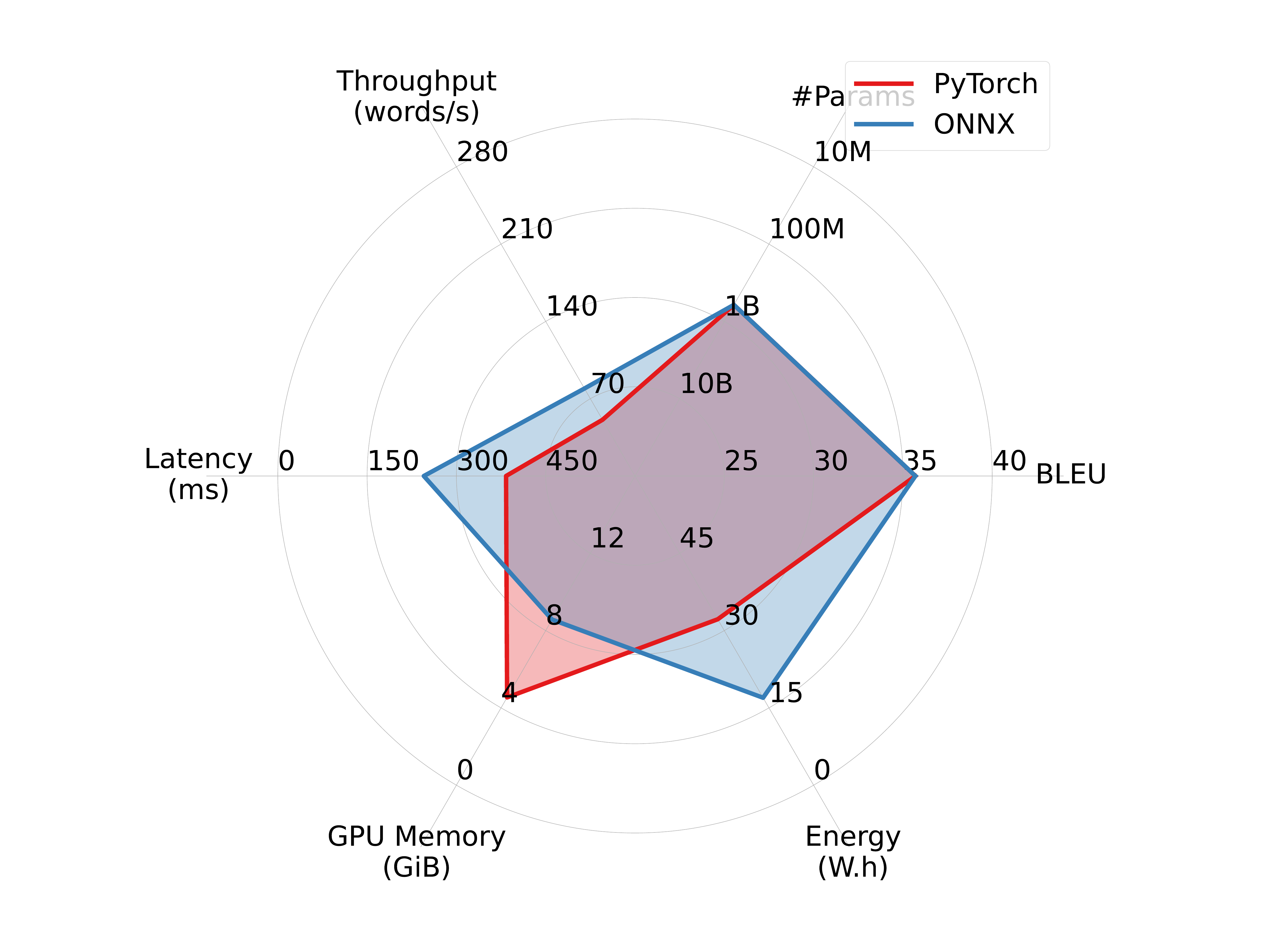}
\caption{MBART M2O. }
\label{fig:wmt14-de-en-mbart-m2o}
\end{subfigure}
\hfill
\begin{subfigure}[t]{0.48\textwidth}
\includegraphics[width=\textwidth, trim={9.4cm 3.5cm 9.5cm 3.5cm},clip]{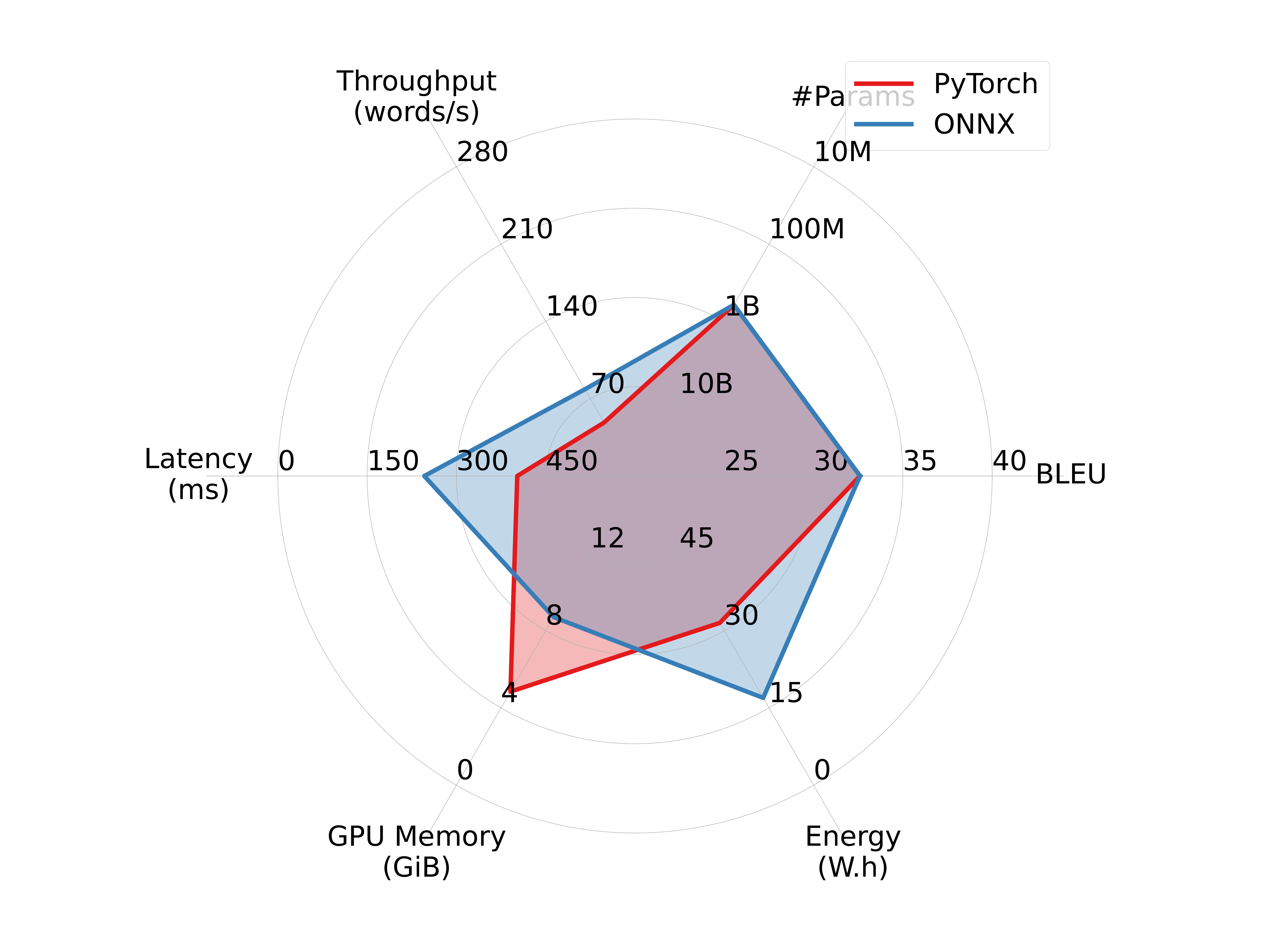}
\caption{MBART M2M. }
\label{fig:wmt14-de-en-mbart-m2m}
\end{subfigure}
~
\begin{subfigure}[t]{0.48\textwidth}
\includegraphics[width=\textwidth, trim={9.4cm 3.5cm 9.5cm 3.5cm},clip]{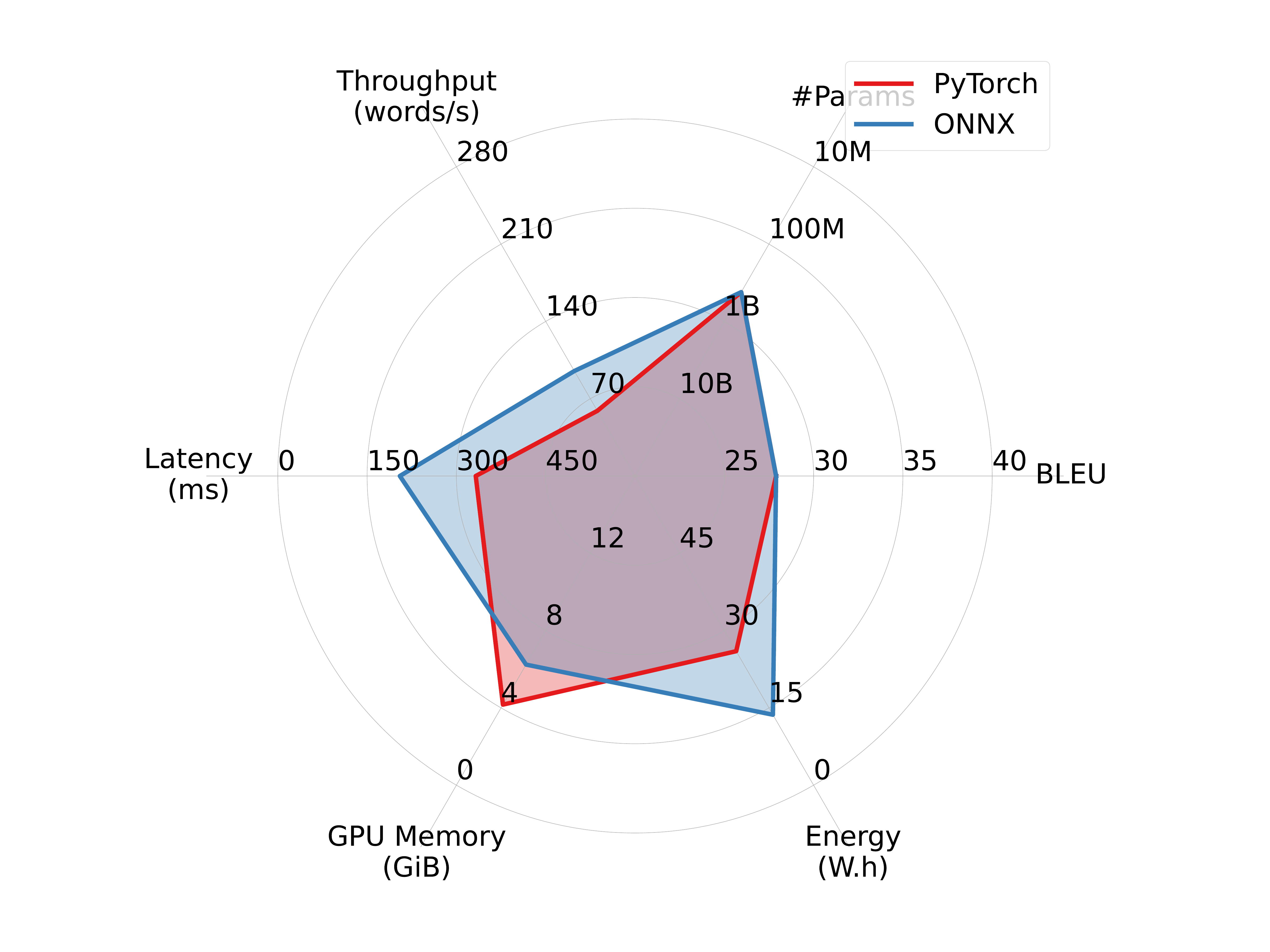}
\caption{M2M100 418M. }
\label{fig:wmt14-de-en-m2m-418m}
\end{subfigure}
\hfill
\begin{subfigure}[t]{0.48\textwidth}
\includegraphics[width=\textwidth, trim={9.4cm 3.5cm 9.5cm 3.5cm},clip]{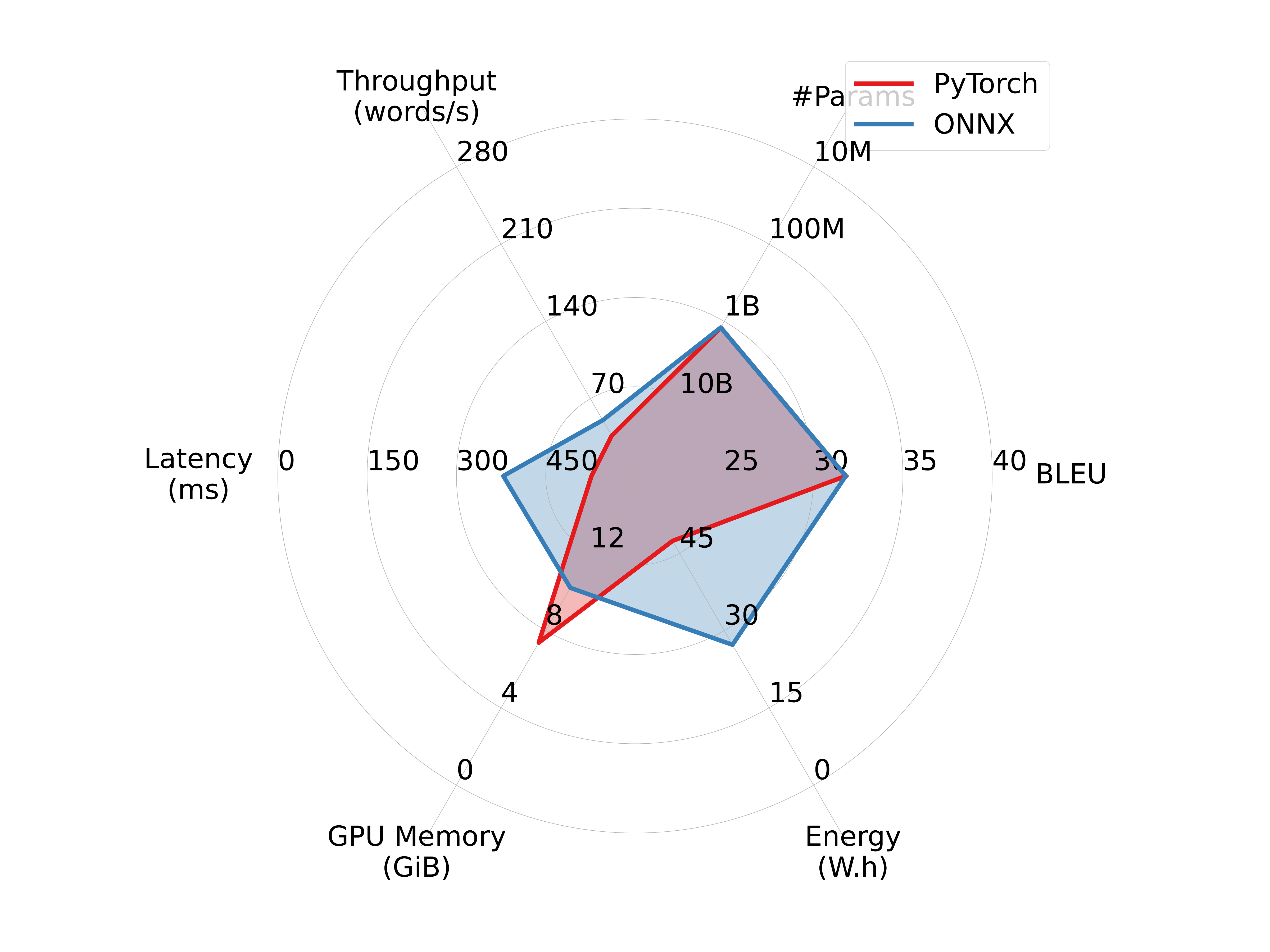}
\caption{M2M100 1.2B. }
\label{fig:wmt14-de-en-m2m-1.2b}
\end{subfigure}
~
\begin{subfigure}[t]{0.48\textwidth}
\includegraphics[width=\textwidth, trim={9.4cm 3.5cm 9.5cm 3.5cm},clip]{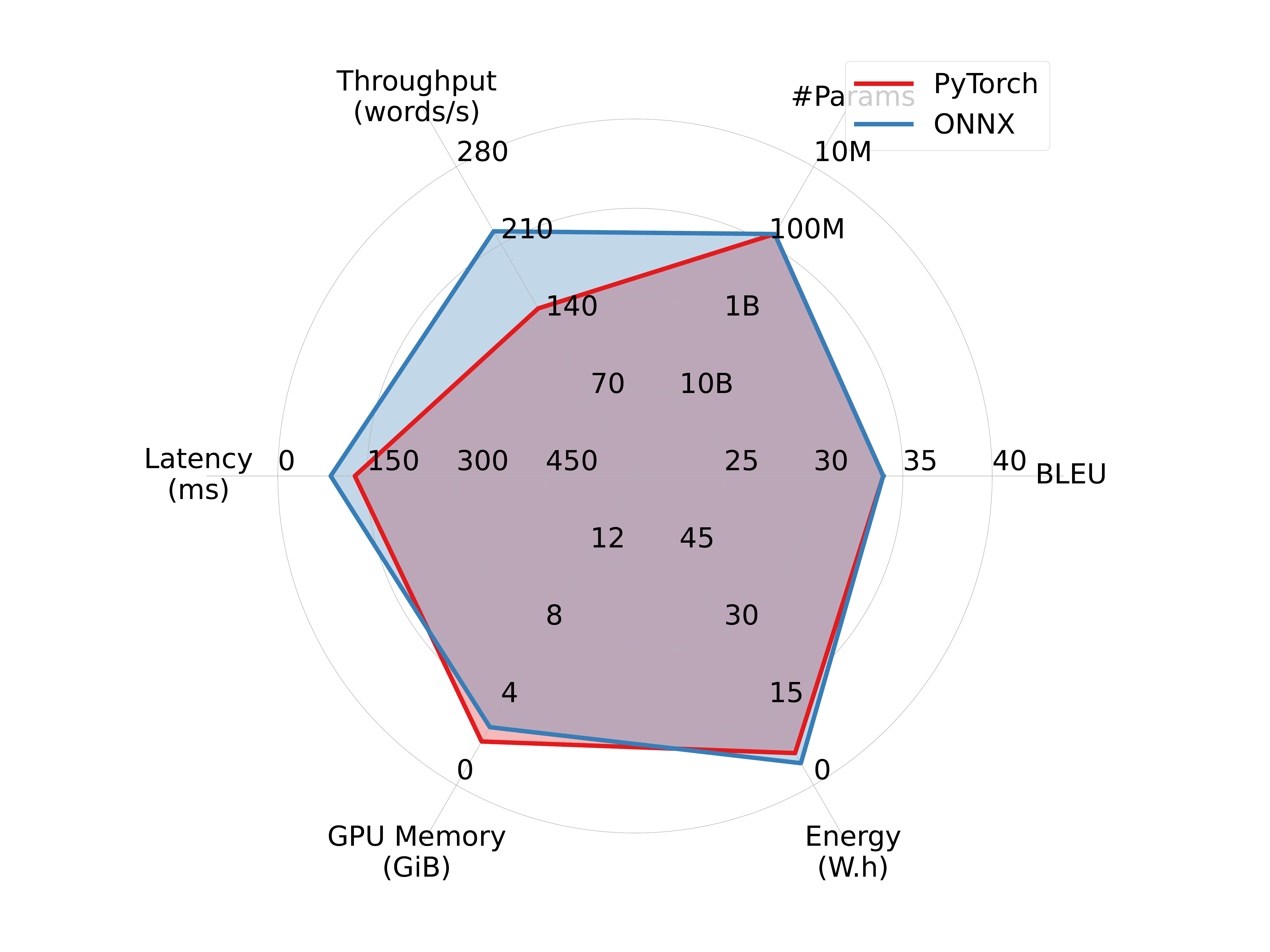}
\caption{OPUS. }
\label{fig:wmt14-de-en-wmt19-meta}
\end{subfigure}
\hfill
\caption{Accuracy and efficiency performance comparisons of five different models while using PyTorch (red) and ONNX (blue) runtime. WMT19 Meta and WMT21 Meta rely on the Fully Sharded Data Parallel (FSDP;~\citealp{fsdp}) in their implementation, which complicates their conversions to ONNX, and are therefore not included in this figure. 
All efficiency metrics are measured in the single-stream scenario; in preliminary experiments, we observe that the efficiency gains from ONNX are marginal in other scenarios, as expected. 
}
\label{fig:onnx}
\end{figure*}

As shown in Figure~\ref{fig:onnx}, when comparing five different models in a single-stream scenario using PyTorch and ONNX runtime, ONNX delivers substantial improvements in throughput, latency, and energy overhead, especially for larger models. However, this comes with an increase in GPU memory consumption, which is likely due to the storage of pre-compiled computational graphs on the GPU. 
WMT19 Meta and WMT21, which utilize the Fully Sharded Data Parallel technique (FSDP;~\citealp{fsdp}), are excluded from this experiment due to compatibility challenges with ONNX and FSDP.

Our preliminary experiments find that ONNX brings marginal efficiency improvements in other scenarios that use larger batch sizes, which is consistent with the observation by \citet{fernandez2023framework}.

\noindent\textbf{Results on WMT14 EN->DE.}

\begin{figure*}[t]
\centering
\begin{subfigure}[t]{0.48\textwidth}
\includegraphics[width=\textwidth, trim={9.4cm 3.5cm 9.5cm 3.5cm},clip]{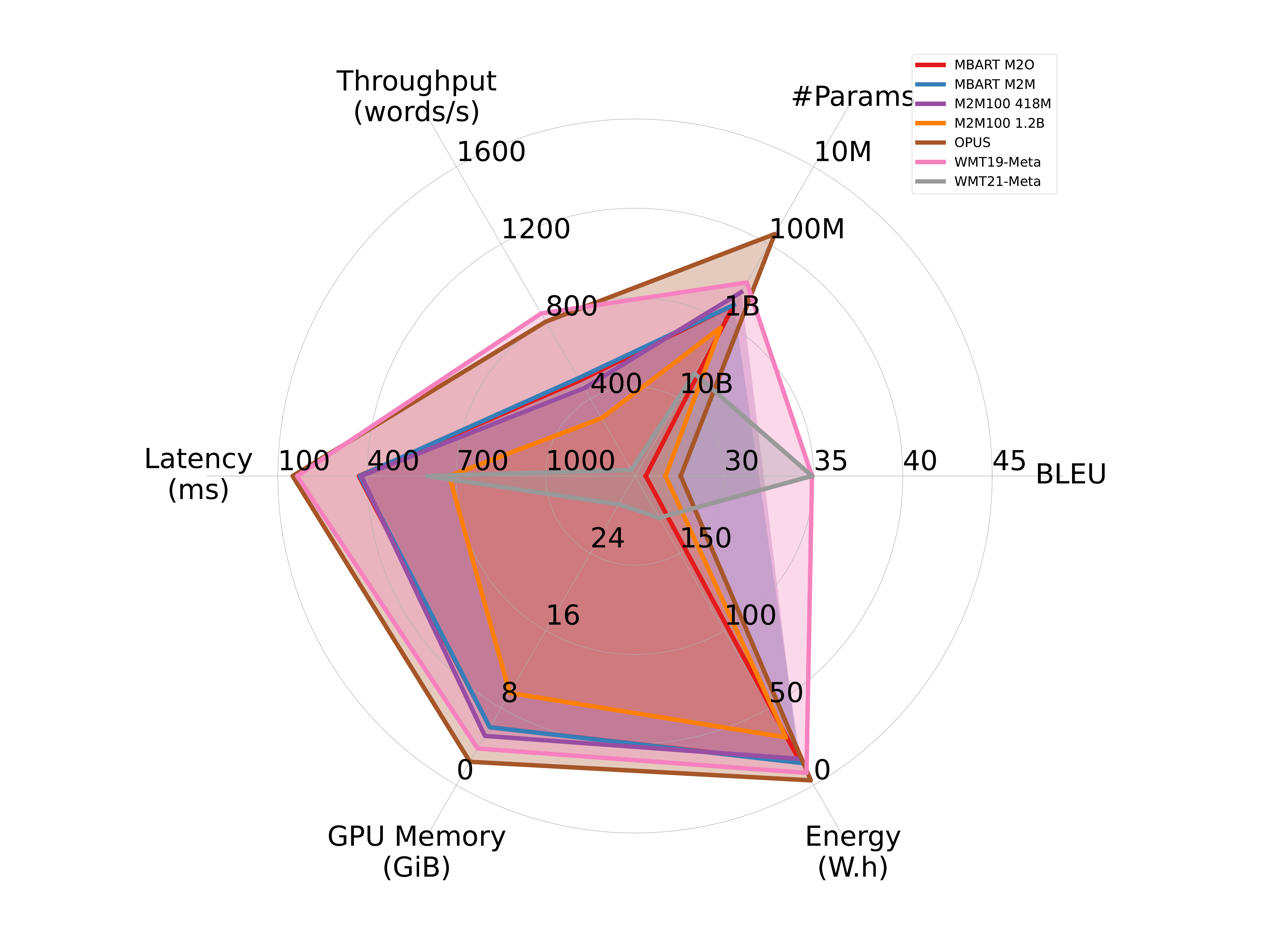}
\caption{FP32. }
\label{fig:wmt14-ende-fp32}
\end{subfigure}
\hfill
\begin{subfigure}[t]{0.48\textwidth}
\includegraphics[width=\textwidth, trim={9.4cm 3.5cm 9.5cm 3.5cm},clip]{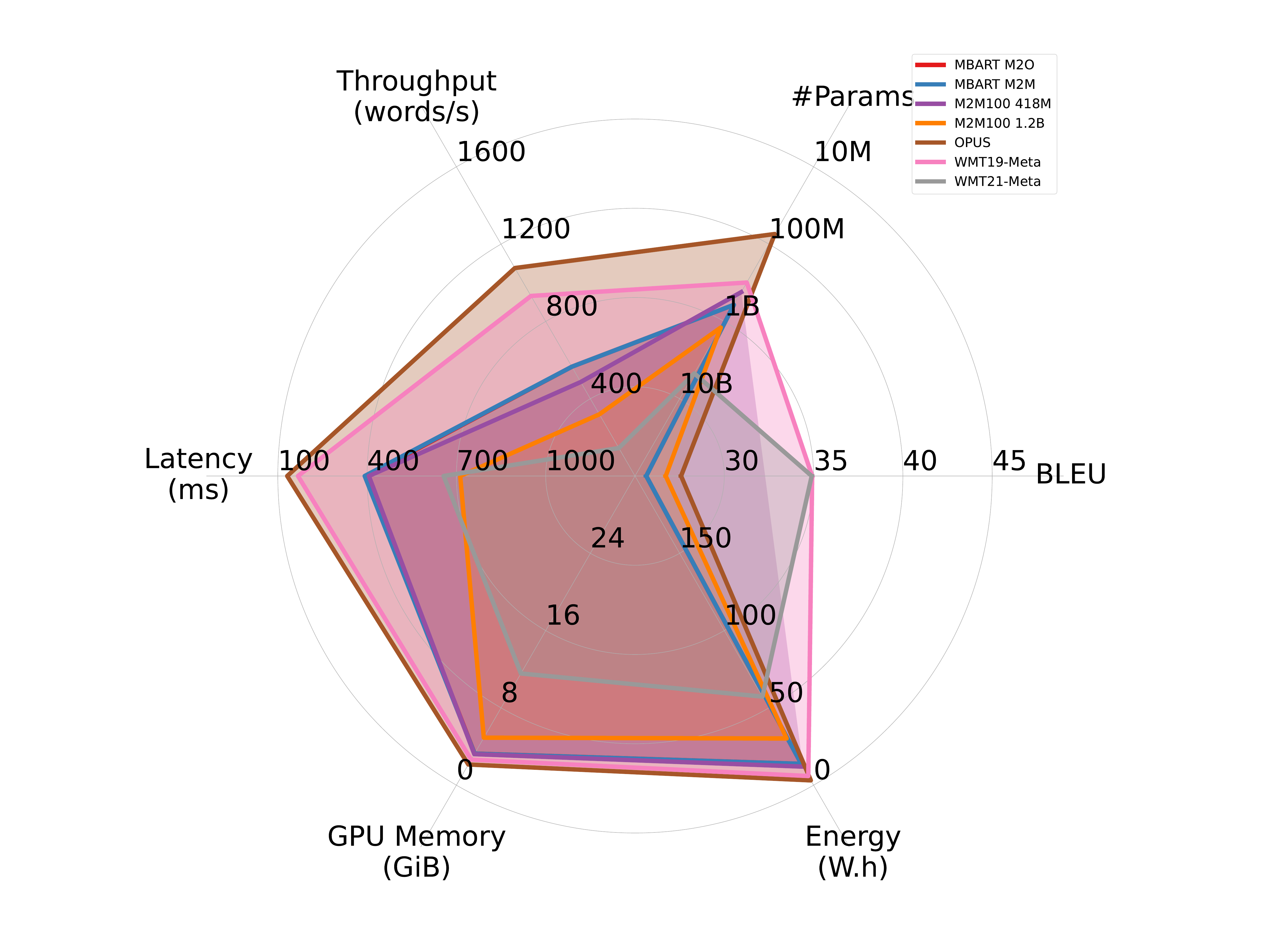}
\caption{FP16 }
\label{fig:wmt14-ende-fp16}
\end{subfigure}
\caption{Performance of various models on the WMT14 EN-DE. Following Figure~\ref{fig:radar}, the figures include throughput metrics from the offline scenario, latency and GPU memory metrics from the single stream scenario, and energy metrics from the fixed batching scenario. For all metrics, {\bf outer rings indicate better performance}. \#Params is presented on a logarithmic scale.}
\label{fig:wmt14-ende}
\end{figure*}

Figure~\ref{fig:wmt14-ende} provides a summary of the efficiency performance of various models on the WMT14 English-to-German (EN->DE) translation task. The results are shown for both FP32 and FP16 models. The observed trends align with those discussed in Section~\ref{sec:experiments}.

\end{appendices}